\documentclass[fleqn,10pt]{article}
\usepackage[margin=1in]{geometry}
\usepackage{authblk}
\usepackage{booktabs}
\usepackage{amsmath}
\usepackage{url}
\usepackage{graphicx}
\usepackage{ntheorem}
\usepackage{hyperref}
\usepackage{appendix}
\usepackage{array}
\usepackage{longtable}
\usepackage{enumitem}
\usepackage{titlesec}
\usepackage{placeins}
\theoremseparator{:}

\usepackage[english]{isodate}

\title{Learning from sanctioned government suppliers: A machine learning and network science approach to detecting fraud and corruption in Mexico}

\author[1, *]{Martí Medina-Hernández}
\author[1, +]{János Kertész}
\author[2, +]{Mihály Fazekas}

\affil[1]{Department of Network and Data Science, Central European University, Vienna, Austria}
\affil[2]{Department of Public Policy, Central European University, Vienna, Austria}
\affil[*]{Corresponding author: medina\_marti@phd.ceu.edu}
\affil[+]{These authors contributed equally to this work}

\date{}
\begin{document}

\flushbottom
\maketitle

\abstract{
Detecting fraud and corruption in public procurement remains a major challenge for governments worldwide. Most research to-date builds on domain-knowledge-based corruption risk indicators of individual contract-level features and some also analyzes contracting network patterns. A critical barrier for supervised machine learning is the absence of confirmed non-corrupt (negative) examples, which makes conventional machine learning inappropriate for this task. 
Using publicly available data on federally funded procurement in Mexico and company sanction records, this study implements positive–unlabeled (PU) learning algorithms that integrate domain-knowledge-based red flags with network-derived features to identify likely corrupt and fraudulent contracts. The best-performing PU model on average captures 32\% more known positives and performs on average 2.3 better than random guessing, substantially outperforming approaches based solely on traditional red flags. The analysis of the Shapley Additive Explanations reveals that network-derived features—particularly those associated with contracts in the network core or suppliers with high eigenvector centrality—are the most important. Traditional red flags further enhance model performance in line with expectations, albeit mainly for contracts awarded through competitive tenders. This methodology can support law enforcement in Mexico, and it can be adapted to other national contexts too.
}

\noindent \textit{Keywords:} Fraud, corruption, measurement, public procurement, Mexico, positive - unlabeled learning


\section*{Introduction}\label{introduction}

The objective of this research is to develop a methodology that identifies and ranks contracts suspicious of fraud and corruption in public procurement (PP). This approach integrates three research strands: corruption risk indicators (``red flags'') \cite{fazekas_objective_2016,instituto_mexicano_para_la_competitividad_anexo_2019}, relational-based network information \cite{fazekas_corruption_2020,czibikNetworkedCorruptionRisks2021}, and positive-unlabeled (PU) learning \cite{jaskiePositiveUnlabeledLearning2022} using real-world examples --- i.e., government sanctions as indicators of fraud and corruption.

\textbf{Public procurement} refers to the process through which government agencies or state-owned enterprises purchase goods and services from private suppliers \cite{oecd_public_nodate}. According to the OECD, PP represents about 12\% to 29\% of total government expenditure in its member countries \cite{oecdPreventingCorruptionPublic2016,oecd_integrity_2007}, an activity highly vulnerable to corruption \cite{oecdPreventingCorruptionPublic2016}. Transparency International estimates that between 10\% and 15\% of PP spending is lost to corruption, though the real cost is likely higher \cite{wiehen_handbook_2006}. Corruption in public procurement is particularly difficult to detect, monitor and prevent because the process is diverse, complex, information is highly asymmetric, and collusion between officials and bidders hides manipulation behind procedures that appear formally legal \cite{oecd_integrity_2007}.

The development of corruption risk indicators or ``red flags'' capable of identifying suspicious contracts has been one of the main objectives in recent PP research. A major contribution in this direction was made by \cite{fazekas_objective_2016}, where the Corruption Risk Index (CRI), a measure based on PP processes and contract characteristics was introduced. The CRI incorporates features known as \textbf{corruption risk indicators} (or ``red flags'') that indicate corruption risks at the individual contract level, such as the presence of a single bidder, absence of a published tender call, a non-open procedure type, and a very short advertisement period, among others. Although the CRI effectively captures integrity risks embedded in specific contracts, it offers limited insight into how corruption emerges through the relationships among the actors involved. 

To overcome this limitation, some researchers have increasingly applied \textbf{network science} tools to analyze corruption from a relational perspective \cite{wachs_corruption_2021, lyraFraudCorruptionCollusion2022}. Typically, PP contracts are modeled as bipartite networks composed of two groups of nodes ---government agencies (buyers) and companies (suppliers)--- connected when a bid is submitted or a contract is awarded \cite{fazekas_corruption_2016,fazekas_corruption_2020,wachs_corruption_2021,czibikNetworkedCorruptionRisks2021,nicolas-carlockOrganizedCrimeBehavior2023}. In other cases, the two groups represent companies and contracts \cite{wachsNetworkApproachCartel2019}, a structure used to detect cartel formation in bidding markets.

Although centralization and clustering in procurement networks are generally associated with corruption risk, their effects vary across countries and in magnitude. \cite{wachs_corruption_2021} found that corruption risk can concentrate either at the network’s center or among peripheral actors, depending on the country. Similarly, \cite{fazekas_corruption_2016} linked distinct clusters in PP networks to different levels of corruption and state capture, while \cite{fazekas_corruption_2020} showed that high-risk agencies often have sparser network neighborhoods than expected. Other network measures, such as diameter, average path length, and average degree, have proven less informative for detecting corruption \cite{luna-pla_corruption_2020}.

For both research approaches -- domain-knowledge-based red flags and network science -- establishing measurement validity and investigative usefulness have been a challenge due to the lack of proven cases of corruption as well as non-corruption, to validate against. In other domains, where learning data is more readily available, \textbf{supervised machine learning} techniques have been quite successful, such as online auction sites \cite{almendra_finding_2013} or shill bidding detection (when participants place fake bids to artificially raise prices) \cite{ganguly_online_2018}.  
In the few cases when labeled data has been used in PP, two major challenges remain. First, the labels employed are often not actual instances of corruption but rather likely cases or grey cases \cite{ferwerda_corruption_2017} or sometimes even using proxy indicators to flag cases . One solution to this problem could be the use of debarments, sanctions or administrative penalties issued by national authorities as positive labels in machine learning models. Although such information is publicly available in some countries \cite{proact_procurement_nodate}, its use has been a largely underexplored in PP research. Importantly, public authorities issuing sanctions may themselves be corrupt or under political control so the derived labels may be biased requiring a careful assessment of data quality. Moreover, sanctions are all but one of many types of hard evidence of fraud and corruption with further examples including corruption convictions against company owners \cite{decarolisCorruptionRedFlags2022}. Using  company sanctions as PP labels in machine learning models face three practical challenges: (1) mapping company-level sanctions to contract-level labels, (2) the absence of negative observations, and (3) data imbalance. Typically, as in this study, data on sanctioned companies do not include the specific contracts involved but only general information about the sanction. Therefore, it is necessary to define criteria for assigning labels to individual contracts, affecting both the training sample and its representativeness. Companies with many contracts will receive more labels, biasing model outcomes. Moreover, using sanctions as positive labels implies the lack of true negative examples, violating the assumptions of traditional Positive-Negative (PN) learning and distorting standard performance metrics. In addition, because only a small fraction of firms are sanctioned, any PP dataset will contain very few positive cases.

The second challenge for applying supervised machine learning to fraud and corruption detection is methodological. The work of Aldana et al. \cite{aldana_machine_2022} represents an important advance in this respect. However, the aforementioned challenges are not fully met. Although labeling all contracts from a sanctioned company as corrupt is plausible, it can overrepresent large companies and risks data leakage without a stratified company-level train-test split. Furthermore, the assumption that all non-labeled companies are non-corrupt treats the problem as PN learning, when in reality only positive labels exist, while in reality the problem represents an example of the \textbf{Positive-Unlabeled (PU) learning} challenge and should be treated accordingly. The same applies to PP machine learning research using criminal convictions \cite{decarolisCorruptionRedFlags2022} or political connections \cite{titlIdentifyingPoliticallyConnected2024}. Like traditional classification, PU learning aims to distinguish positive from negative cases, but without explicit negative labels \cite{jaskiePositiveUnlabeledLearning2022}. This limitation is nontrivial, as the predictive performance of PU classifiers strongly depends on label availability \cite{luo_towards_2023,saundersEvaluatingPredictivePerformance2022}. There is no standard method for PU learning, though several strategies have been proposed; comprehensive reviews can be found in \cite{jaskiePositiveUnlabeledLearning2022,jaskie_positive_2019}. Most have been tested on balanced datasets, while only a few target imbalanced ones \cite{ortegavazquezHellingerDistanceDecision2023}. We focus on two algorithms specifically designed for this scenario: the Positive-Unlabeled Bagging (PU Bagging) \cite{mordeletBaggingSVMLearn2014} and the Hellinger Distance Stratified Random Forest (HDSRF) \cite{ortegavazquezHellingerDistanceDecision2023}.

\textbf{PU Bagging} assigns misclassification costs by aggregating classifiers trained to distinguish positive from unlabeled samples and averaging their predictions. Training uses bagging, a resampling strategy that draws bootstrap samples from the unlabeled data while including all labeled positives. This method handles imbalanced datasets naturally and can be combined with various classifiers, most often Support Vector Machines (SVMs) \cite{mordeletBaggingSVMLearn2014, ortegavazquezHellingerDistanceDecision2023}.

A simpler yet effective alternative is the \textbf{HDSRF} algorithm, which simultaneously addresses PU learning and class imbalance. It employs the Hellinger distance as the split criterion in the random forest’s decision trees, assuming a positive class prior larger than the observed positive proportion. In addition to bagging and random feature selection, the algorithm ensures that all positive samples appear in each bootstrap iteration (hence, ``stratified''). The Hellinger distance has been shown to outperform Gini and Entropy for imbalanced datasets, though it requires knowing the number of positive samples ---a limitation that can be mitigated by treating the class prior as a tunable hyperparameter \cite{ortegavazquezHellingerDistanceDecision2023}.

Both HDSRF and PU Bagging assume that positive examples are Selected Completely at Random (SCAR), meaning the observed positives are a random subset of all positives in an unobserved PN dataset. This assumption is rarely valid and remains one of the key challenges in PU learning \cite{jaskiePositiveUnlabeledLearning2022} and it is also a problem in our case.

Earlier work has made substantial progress in identifying corruption risks using a diversity of approaches. We advance this literature in a number of ways: First, we show that widely available data on suppliers' sanctions can serve as labels to infer risk factors from. Second, we implement a robust and replicable method, tailored to the typical prediction task expected in other countries' PP datasets. This recognizes the PU structure and imbalanced nature of the data. Third, this paper combines, confirms and refines red flags from prior literature such as direct awards; while it also incorporates a wide array of network features derived from the contracting network. Importantly, established red flags are most precise when aggregated to the supplier level, rather than on the contract-level. We also introduced a method for evaluating PU-learning tasks that is useful for ranking instances and robust to cases where predicted probabilities are similar. Finally, our results also demonstrate practical usefulness for detecting fraud and corruption in real-life tenders.

\section*{Data and Methodology}\label{data-methodology}

\subsection*{Datasets}\label{datasets}

The \textbf{contracts dataset} includes 2,301,278 contracts awarded by Mexican federal, state, and municipal governments between 2011 and 2022 using federal funds. It contains contract-level information for 5,304 government entities and 259,534 private suppliers. The dataset was compiled from contract- and procedure-level data available on the Mexican government's CompraNet platform \cite{compranet_datos_nodate}, following \cite{fazekasGlobalContractlevelPublic2024, falcon-cortes_practices_2022}. Full details on the original datasets and the matching process are provided in Supplementary Information \ref{dataset-construction}.

To generate labels for our model, we created the \textbf{sanctions dataset}, comprising 14,535 unique companies. This dataset combines two sources: 12,396 companies identified for issuing invoices for simulated operations (``Empresa que Factura Operaciones Simuladas" or EFOS) \cite{servicio_de_administracion_tributaria_listado_2023} and 2,139 companies sanctioned for involvement in corrupt activities in public contracting (``Proveedores y Constratistas Sancionados" or PCS) \cite{secretaria_de_la_funcion_publica_proveedores_nodate}. Of these, 1,673 companies had contracts in the contracts dataset, with 748 from EFOS and 925 from PCS. More details on the data sources and datasets can be found in Supplementary Information \ref{dataset-construction}. Crucially for the trustworthiness and reliability of our labels, companies sanctioned in the EFOS and PCS datasets are determined by the Tax Administration Service (``Servicio de Administración Tributaria", in Spanish) and the Ministry of Public Administration (``Secretaría de la Función Pública") respectively, formally independent from both the contracting authorities and suppliers in the dataset. We consider potential political bias in the labels as relatively contained, because our data spans across two different federal governments, hence a new administration could investigate wrongdoing by the predecessor. Nevertheless, administrative biases may remain, due to the different auditability or the difficulty of investigation of different procurement procedures. This is a point we will return to when discussing our detailed results.

We treated all contracts of sanctioned companies as fraudulent, regardless of when the sanction was made or the contract signed. This strong assumption implies that sanctions did not have an effect on company behavior, and that contracts of the same company show strong similarity across time. This decision followed a careful analysis of the sanctions' impact on company behavior (see Supplementary Information \ref{labelying-hypotheses}).

\begin{figure}[!ht]
\centering
\includegraphics[width=\linewidth]{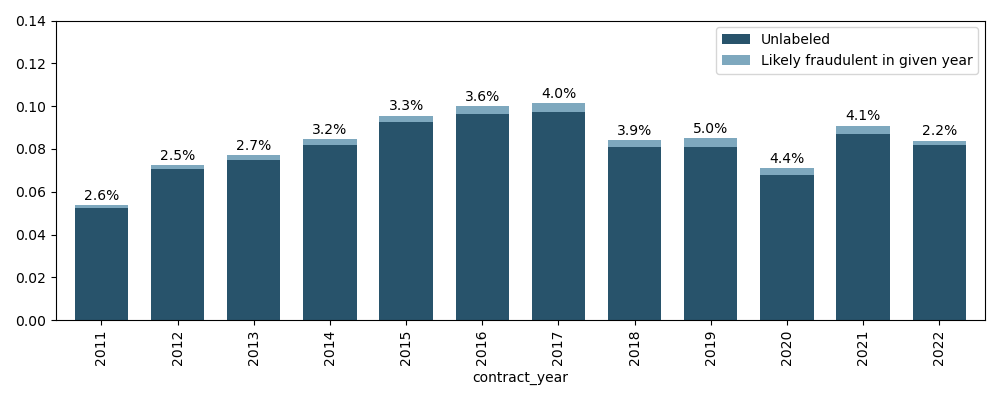}
\caption{Distribution of contracts and labels across years. The height of each bar represents the percentage of contracts (y-axis) in a given year (x-axis) relative to the total number of contracts in the dataset. The dark blue in the bars represent the unlabeled contracts in the dataset, meanwhile the light blue rectangles and the percentages at the top of the distribution indicate the percentage of positive labels in that specific year.}\label{fig:contracts-sanctions-per-year}
\end{figure}

Figure \ref{fig:contracts-sanctions-per-year} shows the yearly distribution of sanctioned contracts after matching with the contracts dataset. After an initial rise in contract numbers, there is a clear decline from 2018, coinciding with the end of the administration of Enrique Peña Nieto (EPN) and the early years of Andres Manuel López Obrador (AMLO). Contract numbers rise again in 2021. The proportion of contracts considered fraudulent (our positive labels) remains between 2.2\% and 5\%.

\subsection*{Feature engineering}\label{only-features}

In line with our research purpose, we included two types of features in the model: domain-knowledge features, which capture well-known indicators of fraud and corruption, and network features, derived directly from the bipartite network and its transformations.

Domain-knowledge features are defined at the contract, buyer, and supplier levels and assigned to each contract in the dataset. Some are directly extracted from the contracts dataset, including contract price, legal framework, supplier size, and supply type. Most features are based on established proxy indicators of corruption, such as Benford's Law, short decision period, non-open procedure type, single bidding, and short submission period which are also combined into a simple composite score the Corruption Risk Index (CRI) \cite{fazekas_objective_2016}. The exact procedure for defining individual red flags and calculating the composite CRI is detailed in Supplementary Information \ref{cri-calculation}. We also included corruption indicators adapted for the Mexican context, inspired in the work of \cite{instituto_mexicano_para_la_competitividad_anexo_2019}, such as suppliers not being recorded in the National Registry and Fragmented Contract Odds\cite{instituto_mexicano_para_la_competitividad_anexo_2019}. Additionally, we included the variables Buyer-supplier active weeks, Buyer-supplier number of contracts per week, Buyer-supplier spending per active Week, which were previously used to analyze corruption in Mexico \cite{falcon-cortes_practices_2022}. Based on the contract-level risk indicators, we also computed aggregated measures, including the average CRI of supplier-buyer edges and CRI of an edge’s neighborhood. Taken together, this set of proxy indicators of corruption in public procurement are comprehensive and offer a detailed picture, even though some key indicators remain imperfect, such as single bidding which suffers from over 85\% missing rate.

Another set of features are derived from network science. Buyer-supplier relationships were represented as a bipartite network, with links between actors connected by a contract and weights equal to the number of such contracts. Variables such as Coreness \cite{wachs_corruption_2021}, Edge Betweenness Centrality, and Competitive Clustering \cite{fazekas_corruption_2020} were calculated. 

We also derived features from the buyer and supplier projections of the bipartite network, including Betweenness, Closeness, Degree, and Eigenvector Centrality. All network centralities were calculated using normalized formulations in the \texttt{igraph} Python package \cite{csardi_igraph_2006}.

All network and domain-knowledge features were determined on an annual basis, such that each value corresponds to a given year. Supplementary Information \ref{features-definition} provides definitions for all 60 features, including domain-knowledge and network-derived features from three network types: the bipartite network and its supplier and buyer projections. Additional characteristics, such as general statistics, distributions, and missing values, are reported in Supplementary Information \ref{dfsummary}.

As described in Supplementary Information \ref{dfsummary}, 12 out of 60 features present some degree of missingness. The missing values of some of them, such as Compliant Submission Period, Tender Period,  Legal Fundament,  are related with the type of procedure: the former two are only defined for non-direct procedures, meanwhile the later is defined only for direct ones.  MAD (Mean Absolute Deviation to Benford's distribution) is calculated only for buyers with at least the equivalent of the quantile 0.75 number of contracts for all buyers in a given year, which is equivalent to a minimum of 42 - 62 contracts, depending on the specific year. Therefore, the contracts with missing MAD are those that belong to buyers with small number of contracts. The missing values of Neighborhood Avg. CRI and Neighborhood Prop. Recorded-Direct Procedures correspond to links with no neighborhoods, i.e. disconnected components in the bipartite buyer-supplier network.  Therefore, the only variables with missing values not created by design, are Legal Framework, Procedure venue, R.F. Procedure Type, R.F. Single Bidder, Supplier Size and Supply Type. Since categorical variables were transformed using one-hot-encoding, we also included a ``missing'' variable for each one of them. In the case of continuous variables, we assigned out-of-range values to indicate missingness for each feature, a common practice in feature engineering \cite{galliPythonFeatureEngineering2020}. 

\subsection*{Train-test split}\label{methods-train-test}

Given the characteristics of our dataset, a simple random-sampling train-test split was not feasible. Our analysis shows that 5\% of suppliers account for nearly 64\% of the contracts, with a Gini coefficient of 0.77 (see Supplementary Information Figure \ref{fig:contract-concentration}). This violates the independent and identically distributed (IID) assumption of many machine learning models, including those relying on bootstrap sampling, such as HDSRF and PU Bagging. These methods draw subsamples of unlabeled observations, which are likely biased toward larger companies, underrepresenting smaller ones. Furthermore, contracts from the same supplier exhibit high similarity (see Supplementary Information \ref{fig:cosine_distance_distribution}), and all contracts of a sanctioned supplier are labeled positive. Consequently, models overrepresenting companies with many contracts would learn to identify large suppliers rather than general patterns of fraud.

To address these issues, we implemented a company-based, undersampled train-test split. Companies in the training set do not appear in the test set, and companies with many contracts are undersampled in training. Details are provided in Supplementary Information \ref{train-test}. On the contrary of other machine learning papers that deal with imbalanced data, we did not balance our test set in order to give a realistic performance of our models in real scenarios, therefore the test set remained untouched.  

We explored two learning approaches to investigate the usefulness of our method: the transductive and inductive learning. The transductive learning objective is to detect observations from an already established sample. The inductive learning objective is to detect observations of data that hasn't been seen. The transductive learning setting would help authorities to detect past contracts, while the inductive setting would be useful for future ones.

Our main objective is framed in transductive learning, aimed at detecting fraud within the current contracts dataset, so the primary train-test split does not consider temporal aspects of the policy of treating fraud in PP, even though the sanctions dataset spans two administrations —EPN and AMLO—which may have applied different strategies.

To evaluate this, we also tested our models on two administration-specific datasets aiming to detect future fraudulent contracts (inductive learning). For each, we trained on the early years of each administration and predicted the second-to-last year of contracts (2017 for EPN, 2021 for AMLO), hiding sanctions from future years in the training set to simulate a real-world scenario where authorities only have past sanctions.

\subsection*{PU Learning Models and validation}\label{models}

Since our dataset is imbalanced and contains positive and unlabeled instances, we used two algorithms suited for the this task: Hellinger Distance Stratified Random Forest (HDSRF) \cite{ortegavazquezHellingerDistanceDecision2023} and PU Bagging with an SVM learner \cite{mordeletBaggingSVMLearn2014}. Both algorithms specifically deal with imbalanced labels \cite{nikpourComprehensiveReviewDatalevel2026}.

Experiments were conducted in Python using the scikit-learn framework \cite{JMLR:v12:pedregosa11a}. Given the relatively recent adoption of PU learning methods, we relied heavily on the publicly available code from \cite{ortegavazquezHellingerDistanceDecision2023}, which provides implementations of several PU learning algorithms, including those used in our study.

We performed hyperparameter tuning and evaluated performance with 4-fold cross-validation. Each fold was stratified using the company-based, undersampled train-test split, ensuring no company appeared in both training and test sets. A fifth set of contracts, not included in the cross-validation folds, was reserved for probability calibration.

\subsection*{Performance evaluation and model selection}\label{performance-evaluation}

One of the main challenges in imbalanced PU learning is model evaluation. Since no negative labels are available and we do not know all positive labels, traditional confusion-matrix-based metrics, such as recall, precision, F-score, AUC-ROC and AUC-PR are not appropriate for model selection. Given that the ultimate objective of our models is to identify a small subset of contracts for authorities to review, we focus on the ranking quality of known positive contracts. As a methodological contribution, we propose tie-aware cumulative sum and lift curves for evaluating our models. These metrics are modified versions of the standard gain and lift curves \cite{berrarPerformanceMeasuresBinary2019}, designed to be robust to predictions with identical probabilities.

Let $\{(y_i, \hat{p}_i)\}_{i=1}^n$ be the set of $n$ true labels $y_i \in \{0,1\}$ and predicted probabilities $\hat{p}_i \in [0,1]$, ordered such that $\hat{p}_1 \ge \hat{p}_2 \ge \cdots \ge \hat{p}_n$. Let $P = \sum_{i=1}^n y_i$ be the total number of known positives, and $\pi = P / n$ the prevalence.

Define the cumulative number of positives up to instance $k$ as
\[
C(k) = \sum_{i=1}^k y_i.
\]

If different instances have equal probabilities there is no unbiased ranking and therefore $C(k)$ cannot give us a strict metric. Let $\mathcal{G} = \{ g_1, g_2, \dots, g_m \}$ denote the partition of indices $\{1, \dots, n\}$ into groups of instances with equal predicted probability. For each $g_j = [a_j, b_j]$, we have $\hat{p}_{a_j} = \hat{p}_{a_j+1} = \dots = \hat{p}_{b_j}$ and $\hat{p}_{b_j} > \hat{p}_{b_{j+1}}$.

The \emph{robust tie-aware cumulative sum} $C_R(k)$ is the instance-based cumulative step function:
\begin{equation}
C_R(k) = 
\begin{cases}
C(a_j), & \text{if } a_j \le k < b_j, \\[4pt]
C(b_j), & \text{if } k = b_j,
\end{cases}
\quad \text{for } k = 1, \dots, n.
\end{equation}

The \emph{null model} for the previous function, assuming a perfect ranking (each instance has a different probability), is the expected cumulative positives under random guessing:
\[
C_0(k) = {\pi}k
\]

The \emph{normalized gain} at instance $k$ is
\[
\text{Gain}(k) = \frac{C_R(k) - C_0(k)}{P},
\]
and the \emph{average gain} over all instances is
\begin{equation}
\overline{\text{Gain}} = \frac{1}{n} \sum_{k=1}^n \text{Gain}(k)
\end{equation}

The lift \cite{berrarPerformanceMeasuresBinary2019} measures how much better the predictions are compared to a baseline, and is defined as the ratio of the probability that the instance is a positive given that the model predicted that it is positive (in our case, $\frac{C_R(k)}{k}$), and the probability that the null model predicts it as positive ($\frac{C_0(k)}{k}$) . Therefore, the \emph{lift curve} at instance $k$ is

\begin{equation}
L_c(k) = \frac{C_R(k)}{C_0(k)}
\quad \text{for } k = 1, \dots, n.
\end{equation}
The \emph{average lift} is 
\begin{equation}
\overline{L_c} = \frac{1}{n} \sum_{k=1}^n L_c(k),
\end{equation}
representing the average factor by which the classifier captures more known positives than random guessing.

Our robust tie-aware cumulative sum $C_R(k)$ is defined as a step function over instances of equal predicted probabilities. The gain at a specific instance is the difference between the classifier’s $C_R(k)$ and the null model $C_0(k)$, while the lift is the ratio between $C_R(k)$ and $C_0(k)$. Average gain measures the proportion of known positives captured beyond random guessing, and average lift measures the factor of improvement over random guessing. Unlike standard average gain and average lift  \cite{berrarPerformanceMeasuresBinary2019}, our method calculates gain only after all instances with the same probability are accounted for, making it robust to classifiers that assign identical probabilities to multiple instances.

This robust tie-aware cumulative sum $C_R(k)$ is conceptually similar to a ``recall curve,'' which plots recall (True Positive Rate) at every classification threshold based on unique predicted probabilities. The key difference is that $C_R(k)$ is defined by instances, not thresholds. This allows us to distinguish between a classifier with a smooth probability distribution and one that assigns blocks of equal probabilities, which is particularly useful for auditing purposes. This approach also prevents the common issue of assigning a recall value of 1 to classifiers that predict all instances as positive. In our modified gain, all instances except the last receive zero gain; therefore, the average gain for a classifier that predicts every instance as positive will be close to zero.

Our selected model is tested for significance using permutation testing \cite{berrarPerformanceMeasuresBinary2019} indicating the probability of observing results as extreme as our model, given that the null hypothesis is true.

\subsection*{Interpretation of the model}\label{methods-interpretation}

Considering potential applications, we present an interpretation of the model in terms of SHapley Additive exPlanations (SHAP values) for tree-based models 
\cite{lundbergLocalExplanationsGlobal2020}, an efficient method for the exact computation of Shapley values \cite{shapleyValueNPersonGames2016}. Shapley values provide a principled way to locally attribute each considered feature’s contribution to an individual machine learning prediction \cite{lundbergUnifiedApproachInterpreting2017, lundbergConsistentIndividualizedFeature2019}.

Specifically, we focus on the average SHAP values of features, which serve as a measure of feature importance for the model. Additionally, we analyzed the distribution of SHAP values for individual features to assess the direction of their influence --that is, whether they contribute positively or negatively to the model’s predictions.

\section*{Results}\label{results}

\subsection*{Performance}\label{results-performance}

\begin{figure}[!ht]
\centering
\includegraphics[width=\linewidth]{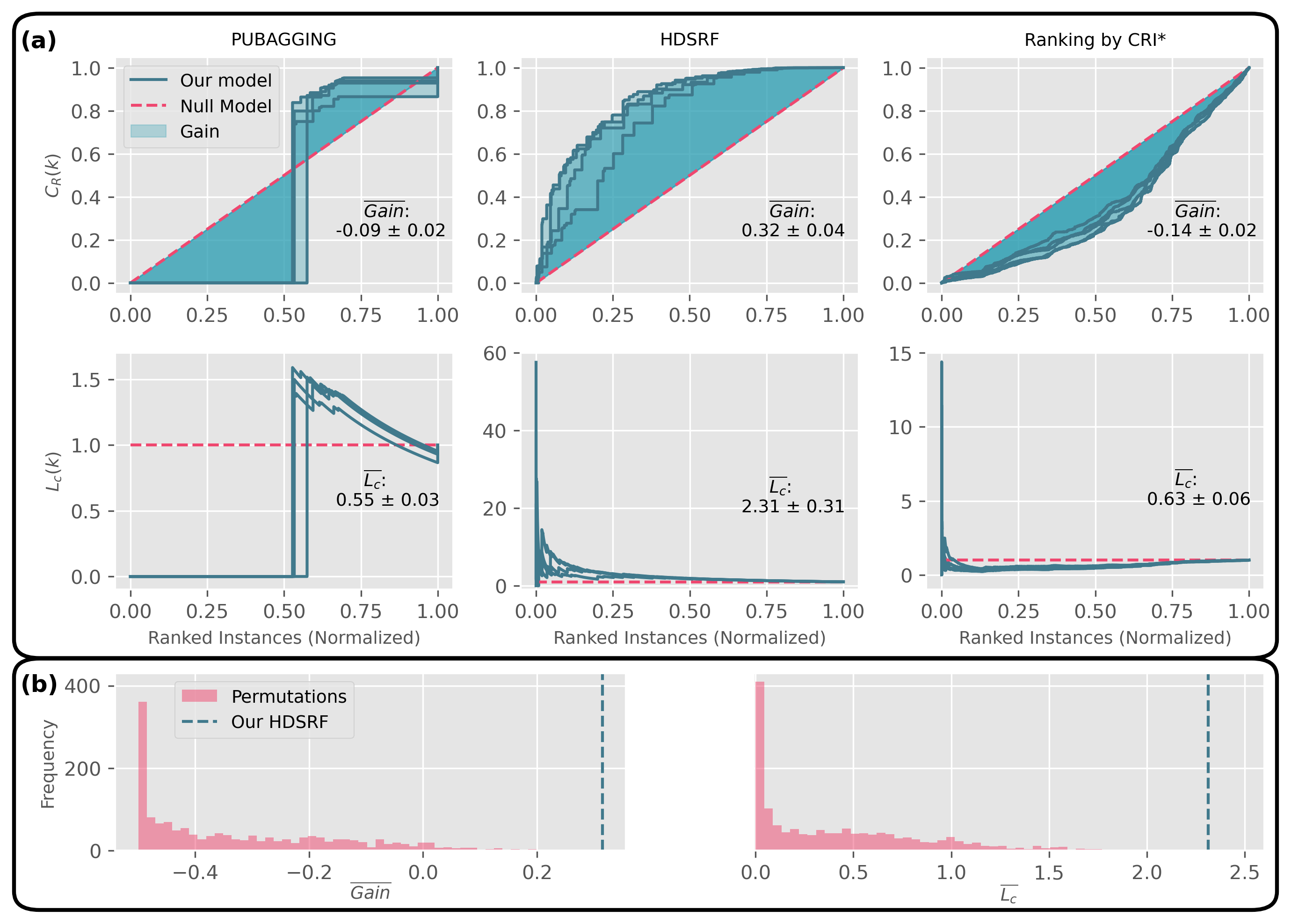}
\caption{Model's performance. In panel (a) each blue line corresponds to the robust cumulative sum ($C_R(k)$) --top row-- or the lift curve ($L_c(K$) --bottom row-- of each of the 4-fold cross-validation sets for the examined models in comparison with a random classifier (red dashed line). The colored area corresponds to the gain: the difference between the robust cumulative sum and the null model. *Ranking by CRI implies to rank the contracts using only the CRI and estimate how well it captures labeled contracts without any machine learning methodology. The figures indicate that the HDSRF outperforms the PU Bagging model for every set. Moreover, the straight line at value zero up to $\sim$ 0.5 normalized ranks in the PU Bagging algorithms shows that around 50\% of the observations in the test set are classified with the highest predicted probability, which makes unfeasible to work for ranking purposes. The figures on panel (b) correspond to the results of the permutation testing for our best HDSRF model. We observe that in both cases, the average gain and average lift of our model overpass all the models created with permuted labels. 
 }\label{fig:fig2}
\end{figure}

The performance of the models can be observed in panel (a) of figure \ref{fig:fig2}. In the top row of the panel we observe the robust cumulative sum ($C_R(k)$), while in the bottom row the lift curve ($L_c(K$) is shown. Each of the blue curves in the figures correspond to each fold of the cross validation. The first two columns of this panel correspond to the PU Bagging and HDSRF models in the transductive setting. Since extensive research has been done in the creation of the ``red flags'', a natural comparison with our models is to rank the instances with CRI \cite{fazekas_objective_2016} and evaluate how different it is from our machine learning models. The comparison can be observed in the most right column of the panel (a). It is clear that HDSRF outperforms PU Bagging as well as ranking by CRI in both the robust cumulative sum and lift curve. On average, HDSRF captures 32\% more known positive instances than random guessing, while PU Bagging and ranking by CRI both considerably underperform this benchmark. Similarly, in the lift curve, HDSRF captures, on average, 2.3 times more known positives than random, compared with 0.55 for PU Bagging and 0.63 of the CRI ranking.

The underperformance of PU Bagging can be explained by its tendency to assign the same probability to many test instances. The sudden rise in gain around 50\% of instances occurs because gain is calculated over groups of instances with equal probability; PU Bagging assigned the highest predicted probability to that many instances (see Supplementary Information \ref{probability-distributions}). This behavior suggests that PU Bagging tends to overclassify instances as positive, limiting its usefulness for ranking. In contrast, HDSRF exhibits a smoother, steadily increasing performance.

One might argue that contract concentration in the test set could inflate performance metrics. To address this, we also evaluated performance on a uniform test set, following the undersampling procedure described before. As shown in Supplementary Information \ref{performance-uniform-testset}, the performance is very similar, indicating that it is independent of contract concentration in the test set.

We calculated the average gain and average lift also for the inductive setting, showing very similar results: for the EPN administration, we got an average gain of -0.05 and 0.31 and an average lift of 0.62 and 2.31 for PU Bagging and HDSRF, respectively. For the AMLO administration, average gain of -0.2 and 0.29 and average lift of and 0.36 and 1.81, also for PU Bagging and HDSRF respectively (see Supplementary Information \ref{performance-inductive}). In both administrations, the HDSRF is significantly better than the PU Bagging, and the performance is very similar to the transductive setting. 

Given that the HDSRF showed the best results for average gain and average lift, we proceeded to test for significance through random permutation tests \cite{berrarPerformanceMeasuresBinary2019}. The results of the test are observed in the first two plots (left to right) of the panel (b) of figure \ref{fig:fig2}. The results indicate that the results of our model are out of the distribution of the permuted labels for average gain and average lift. 
 
\subsection*{Feature Importance}\label{results-feature-importance}

Given the significantly better performance of HDSRF, we focus on this model for feature analysis and interpretation. The mean SHAP values of HDSRF are shown in Figure \ref{fig:general-shapvalues}. Panel (a) presents the top \newcommand{\topkval}{30 } most important features according to their absolute mean SHAP value. After feature engineering, the model includes 69 features, 9 of them resulting from one-hot encoding. Supplier Coreness (weighted degree), Supplier Eigenvector Centrality, and Supplier Proportion of Recorded Direct Procedures are the top three features, highlighting the importance of a supplier's network position and contract composition in detecting fraud.

\begin{figure}[!ht]
\centering
\includegraphics[width=\linewidth]{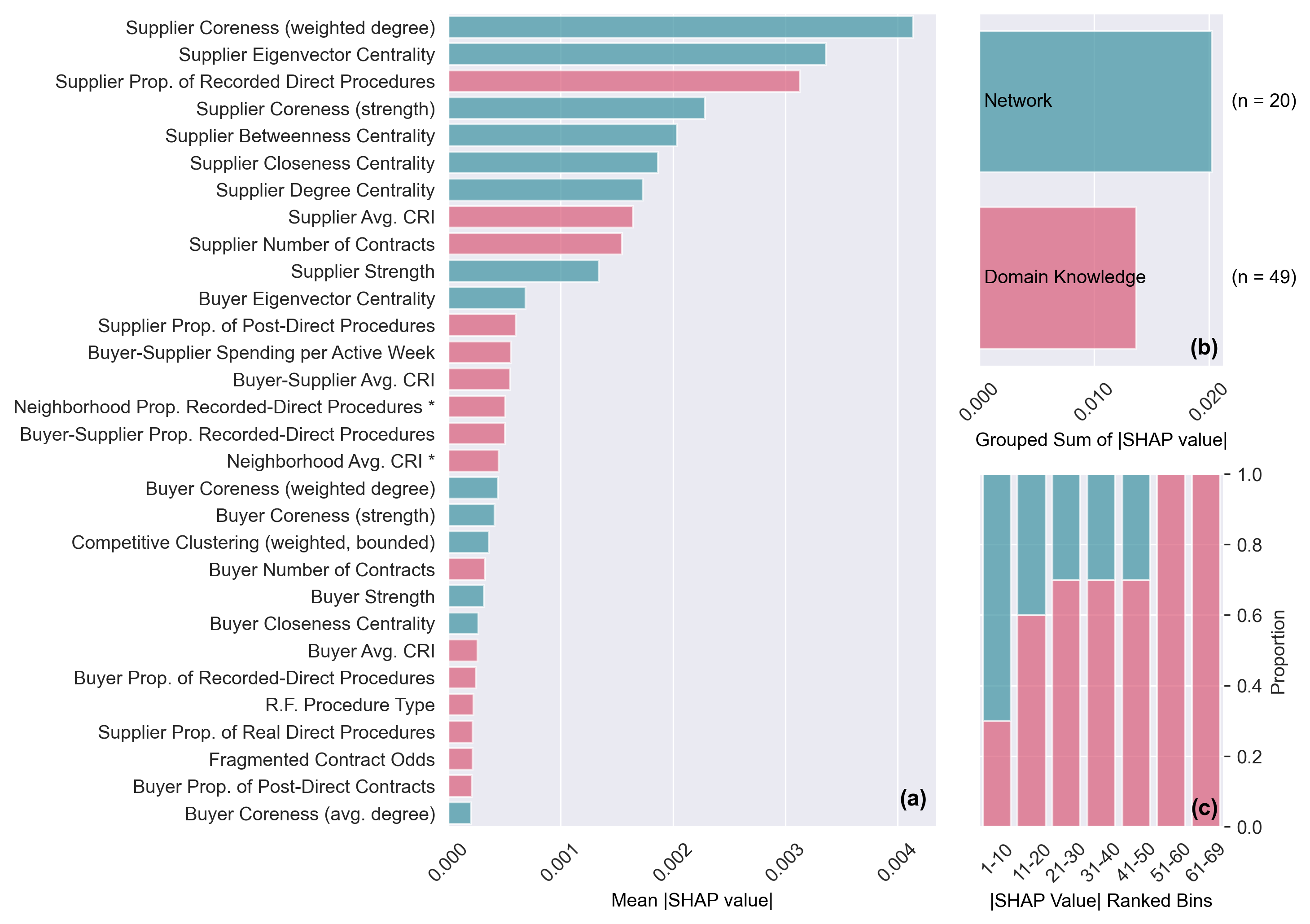}
\caption{Top \topkval most important features of the model. In (a) we observe the top \topkval most important features of our HDSRF according to their mean absolute SHAP value, where the top 10 features are dominated by network features (7 out of 10). The accumulated influence of different variable types is shown in (b). The network features of the model have the highest sum of absolute mean SHAP values. The position of individual features over the entire set of features is depicted in (c). The top positions are a mixture of network and domain-knowledge features, meanwhile the bottom ones are dominated by domain-knowledge features, even though they are present in all of the ranked bins. The definition of each feature can be read in Supplementary Information \ref{features-definition}. * These are features that are take domain-knowledge information by leveraging the network structure}\label{fig:general-shapvalues}
\end{figure}

Panel (b) shows the grouped sum of absolute average SHAP values by feature type. Grouped, network features contribute most to the predicted probability of a contract being corrupt. Although domain-knowledge features are rarely among the very top, taken together they still have a substantial impact on the predicted probability. Among the top \topkval features, domain-knowledge indicators typically feature as company-level aggregates of contract-level measures, such as the proportion of recorded direct and post-direct procedures and average CRI.

Panel (c) provides an overview of feature type rankings across all variables. The first 10 features are a mixture of all types, while most of domain-knowledge dimensions appear at the lower end of the mean SHAP value ranking. 

\begin{figure}[!ht]
\centering
\includegraphics[width=\linewidth]{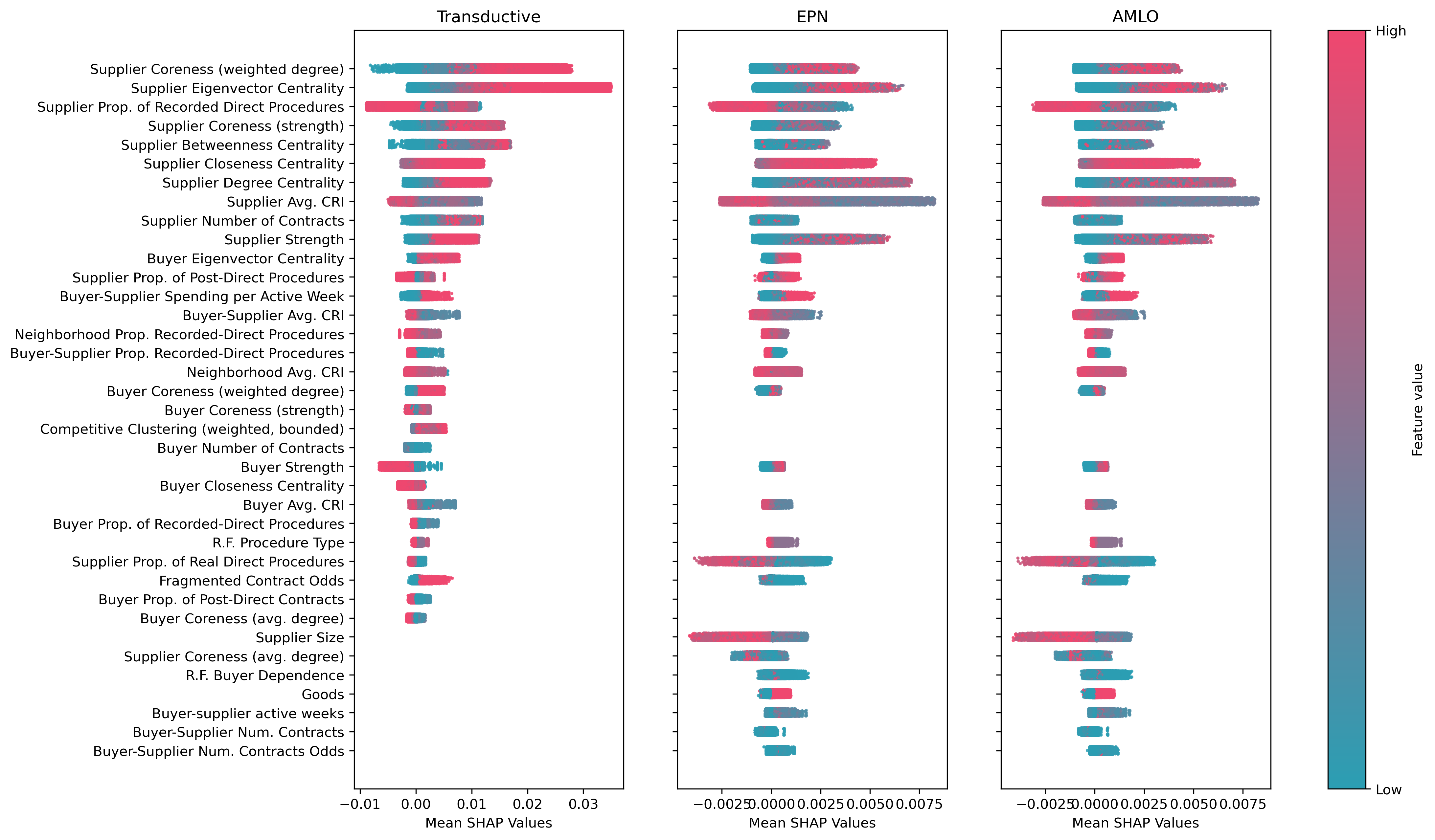}
\caption{Mean SHAP values of top \topkval across transductive and inductive (EPN and AMLO administration) learning. We can observe the non-absolute SHAP values of each feature per instance colored by the values of the feature. All of them shows the average SHAP values of \topkval features ordered by the absolute SHAP value in the transductive setting. The empty spaces show that a feature is not present in the top \topkval for that setting. }\label{fig:YS-comparison}
\end{figure}

We also compared transductive and inductive learning settings across the EPN and AMLO administrations. Figure \ref{fig:YS-comparison} shows non-absolute SHAP values for each feature per instance, colored by feature values. All plots display the average SHAP values of \topkval features, ordered by importance in the transductive setting. Empty spaces indicate that a feature is not present in the top \topkval for that setting.

Several observations arise from this comparison. We see general stability among the top \topkval features across the three settings. First, the three settings share 22 features out of \topkval and the direction of their relationships is largely consistent. Additionally, all features present in the inductive setting are shared across both government administrations. Although the mean SHAP values are generally smaller in the inductive setting compared to the transductive one, the relative impact of each feature is similar. It is worth noting that some highly correlated variables, such as those related to Competitive Clustering, appear in the inductive setting but not in the transductive one, or appear in the transductive but not in the inductive. This suggests that a method based on SHAP values may have difficulty distinguishing the contributions of highly correlated features.

\subsection*{Model Interpretation}\label{results-interpretation}

The most relevant model features were identified based on their absolute mean SHAP values and theoretical importance. Their corresponding SHAP dependence plots were subsequently analyzed (Figures \ref{fig:shapNETWORK} and \ref{fig:domknowledge}). These plots convey four types of information. First, they show the relationship between a feature’s value and its contribution to the predicted probability of an individual instance (SHAP value). Second, the colors indicate their association with other features used in the model. Vertical color changes suggest feature interactions, while horizontal color changes indicate correlations. With interaction, we mean that the effect of feature A on the dependent variable (expressed as SHAP values) changes depending on the values of feature B. Additionally, the grey band centered around zero in the SHAP values denotes the boundaries corresponding to the 10th and 90th quantiles of the SHAP value distribution across all features. This provides an intuitive indication of how common or uncommon specific SHAP values are. These boundaries do not represent statistical significance but rather indicate the relative extremity of SHAP values within the overall distribution. Finally, the blue histogram in the background illustrates the distribution of the feature values.

\subsubsection*{Network features}

\begin{figure}[p]
\centering
\includegraphics[width=\linewidth]{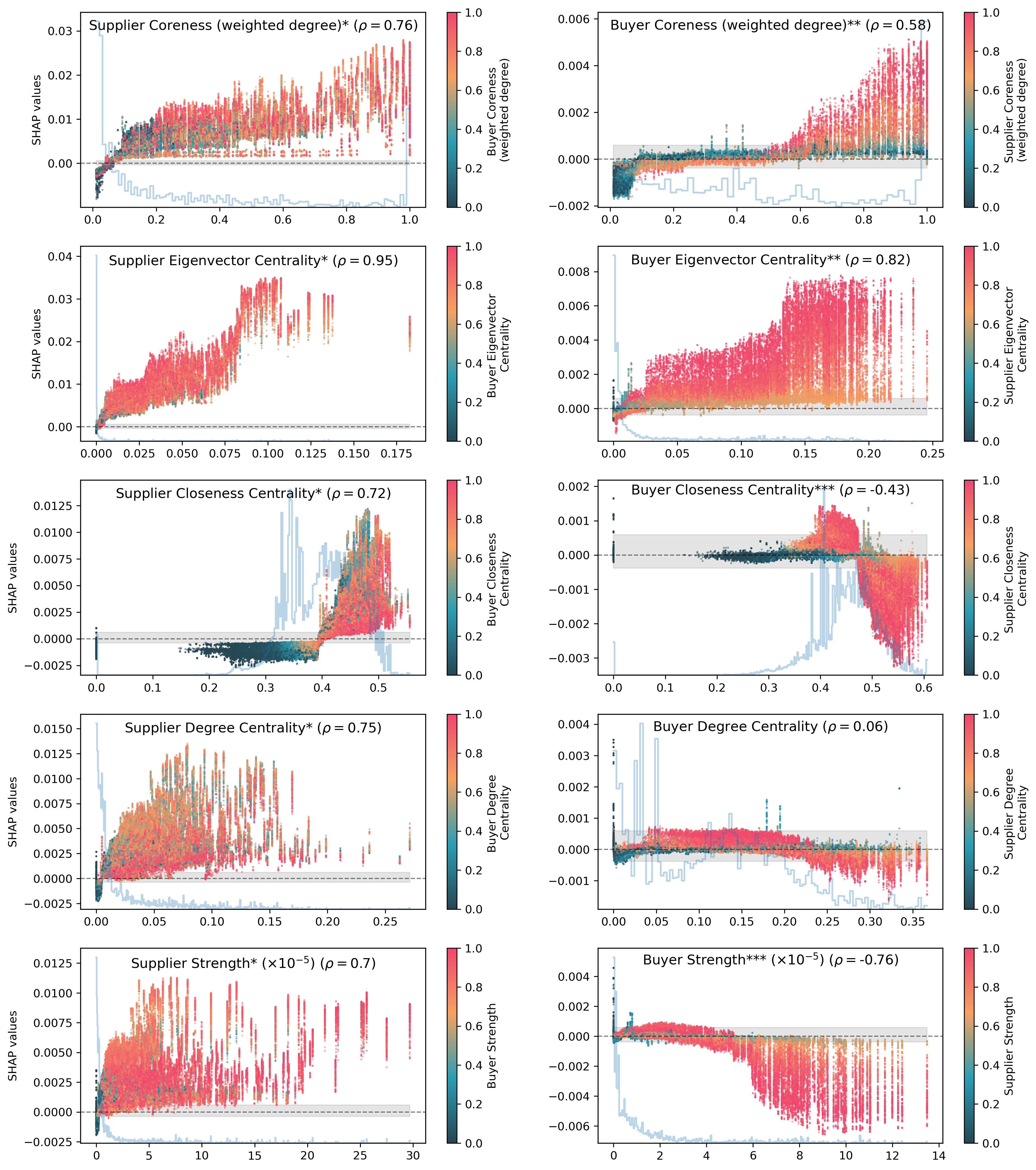}
\caption{Selected network features' SHAP dependence plots. This figure presents the SHAP dependence plots for ten selected network features from the model and its Pearson correlation($\rho$), along with the distribution of the feature in the background (in light blue). The y-axis indicates the features' contribution to the predicted probability of being fraudulent (SHAP value) for each contract, while the x-axis indicates the original values of the feature. The plots in the left column correspond to features related to the supplier, while those on the right correspond to the buyer. Colors correspond to the values of other variables included in the model label. The $*, **, ***$ symbolize the belonging of the feature to the top 10, 20 and 30 most important features of the model,respectively. The grey band around zero represents the 10th and 90th quantiles of the SHAP value distribution across all features, and the blue histogram in the background the distribution of the feature analyzed in the given figure.}\label{fig:shapNETWORK}
\end{figure}

In Figure \ref{fig:shapNETWORK}, we show the SHAP dependence plots of ten network features from the model and the Pearson correlation ($\rho$) between the SHAP values and the original values of the model. These features capture the positions of suppliers and buyers in the bipartite network (for coreness) and in their respective projections of the Mexican public procurement network. Supplier SHAP values are colored by the corresponding buyer features, and vice versa.

Both supplier and buyer \textbf{coreness} show a positive correlation with the SHAP values, meaning that contracts involving actors closer to the bipartite network core receive higher positive contributions to the probability of being fraudulent. For suppliers, this relationship is consistent regardless of the buyer’s position, while for buyers, SHAP values are only high when the contracts are signed with suppliers near the core.

Supporting this observation, supplier and buyer \textbf{eigenvector centralities} also display a positive correlation with SHAP values. In the supplier projection, high supplier eigenvector centrality indicates shared buyers with high-degree suppliers, forming a densely connected supplier core. Similarly, high buyer eigenvector centrality reflects shared suppliers with well-connected buyers. When buyer eigenvector centrality is low but supplier eigenvector centrality is high, the contribution becomes slightly negative, though with small magnitude. 

The similar positive correlation of coreness and eigenvector centrality with their SHAP values, together with a Pearson correlation of 0.68 between supplier coreness and supplier eigenvector centrality, suggests a clear core structure in the network where fraudulent contracts are concentrated. However, as noted by \cite{wachs_corruption_2021}, such a concentration of high-risk contracts in the core is not observed in all countries, indicating that the pattern seen in Mexico is not necessarily expected.

For both \textbf{degree} and \textbf{strength}, supplier features show a positive correlation with SHAP values, while the corresponding buyer features show a negative one. A high degree in the supplier projection indicates a \emph{large number of suppliers sharing common buyers}, whereas high strength indicates a \emph{large number of shared buyers with other suppliers}. This distinction is crucial for interpretation: high supplier strength may reflect coordinated behavior among suppliers to obtain contracts from the same set of buyers. Conversely, in the buyer projection, its negative association with SHAP values may instead indicate recurrent relationships with trusted suppliers or reflect biases in the sanctions dataset.

\textbf{Closeness centrality} shows a similar pattern: supplier closeness is positively correlated with SHAP values, while buyer closeness is negatively related to them. The first association indicates that likely fraudulent contracts are signed with firms that, on average, have fewer buyer intermediaries with other suppliers. This supports the interpretation mentioned beforehand, where both high supplier strength and closeness may reflect coordinated behavior among suppliers to obtain contracts from the same set of buyers. Conversely, in buyer closeness centrality, the pattern suggests that buyers who are structurally closer to others by fewer supplier intermediaries, contribute negatively to the predicted probability of being a fraudulent contract.

Overall, these network features reveal consistent patterns. In all cases, supplier-related features exhibit stronger correlations with their corresponding SHAP values compared to their buyer counterparts. This can be explained by the nature of our labels: since they are derived from company-level sanctions, and contract characteristics tend to be highly similar within the same supplier, the model likely relies more on supplier-related attributes to identify risky contracts. Furthermore, because sanctions are imposed at the company level, the model is better fit to capture contract features that reflect company behavior rather than buyer-side dynamics. Another notable pattern is that all supplier-related features show positive correlations with SHAP values, whereas on the buyer side, this is true only for coreness and eigenvector centrality.

\subsubsection*{Domain-knowledge features}

\begin{figure}[!ht]
\centering
\includegraphics[width=1\textwidth]{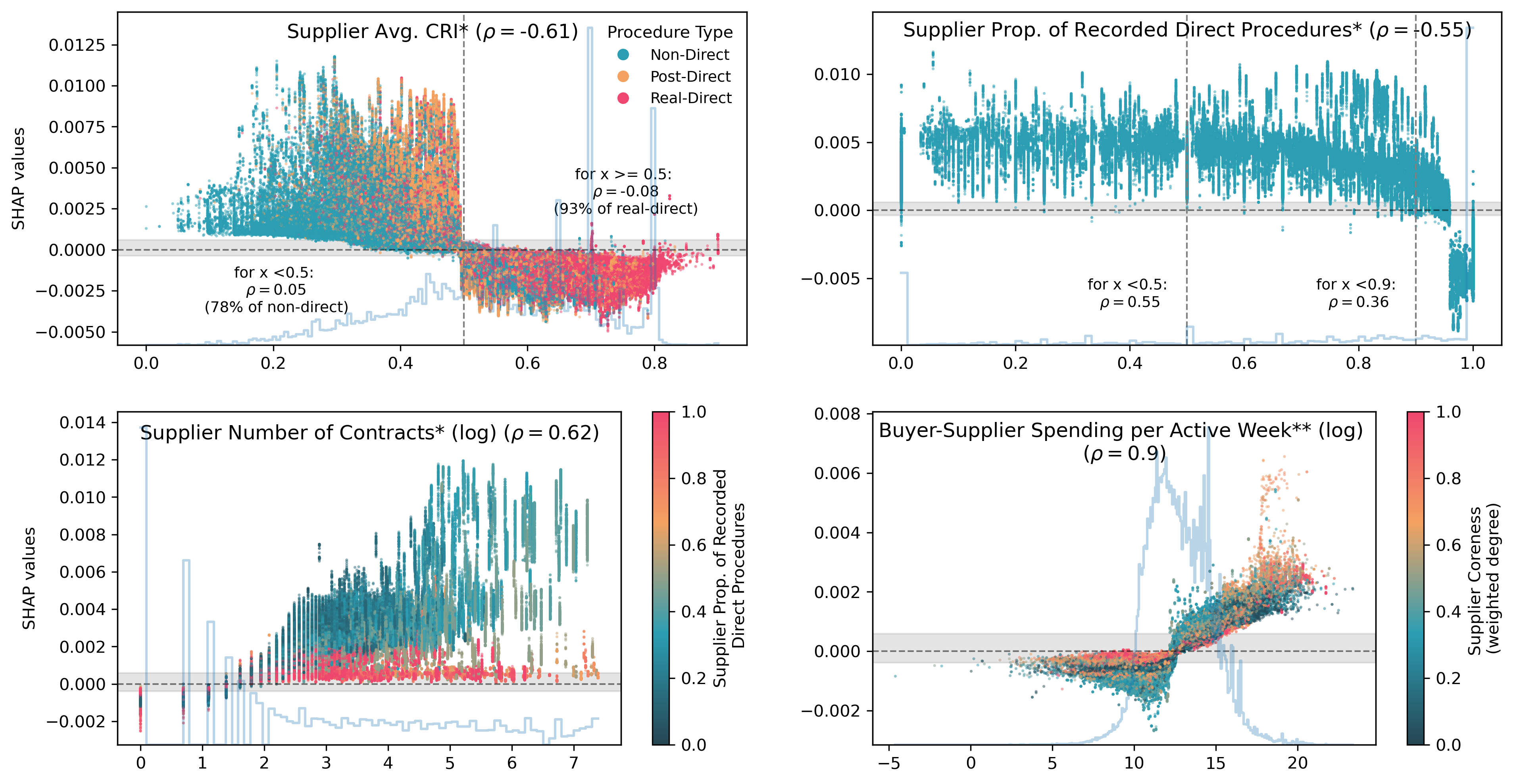}
\caption{Selected Domain-Knowledge features' SHAP dependence plots. This figure shows the SHAP dependence plots of four selected domain-knowledge features of the model, along with the distribution of each feature in the background and the Pearson correlation ($\rho$). Colors correspond to the values of other variables included in the model label. The $*, **, ***$ symbolize the belonging of the feature to the top 10, 20 and 30 most important features of the model, respectively. The grey band around zero represents the 10th and 90th quantiles of the SHAP value distribution across all features, the blue histogram in the background the distribution of the feature analyzed in the given figure, and the dashed vertical lines to specific thresholds of features' values. }\label{fig:domknowledge}
\end{figure}

In Figure \ref{fig:domknowledge}, we present four continuous domain-knowledge features of the model. Two features display a complex and weak relationship with predicted risks (Supplier Prop. of Recorded Direct Procedures and Supplier Avg. CRI), while the two others behave closer to theoretical expectations (Supplier Number of Contracts and Buyer-Supplier Spending per Active Week).  

The complex relationship between SHAP values and both \textbf{Supplier Prop. of Recorded Direct Procedures} and \textbf{Supplier Avg. CRI} hides two different dynamics, centered around the use of direct awards. Although best practices in public procurement recommend avoiding direct procedures –those contracts that are assigned without competition–, in Mexico these account for 74.8\% of all contracts in our dataset. This large subsample in our dataset, however, introduces a likely bias in the labels we use for training, as audits leading to sanctions are more common in non-direct, open procedures, than in direct procedures. This is due to the fact that complex open procedures with extensive audit trail are more readily amenable to audits leading to sanctions rather than simple, direct awards \cite{gerardinoDistortionAuditEvidence2024}. Hence, it is possible that suppliers which receive exclusively or nearly exclusively direct awards are less likely to be sanctioned due to the lack of auditable information. While among suppliers with low-to-medium direct award share, the expected positive relationships between direct awards as a red flag and also the other red flags are present.

Supporting this interpretation, the SHAP dependence plot of Supplier Avg. CRI (Figure \ref{fig:domknowledge}) reveals two distinct regions. For feature values below 0.5, 78\% of non-direct procedures fall in this range and are associated with higher average SHAP values, although the Pearson correlation is close to zero. For values above 0.5, 93\% of real-direct procedures fall in this region, which exhibits consistently low SHAP values and likewise a weak Pearson correlation. The overall negative correlation is therefore driven by the contrast between non-direct and real-direct procedures, while correlations within each region are small. See also Supplementary Information \ref{additional-shap}.

A similar pattern can be found in Supplier Prop. of Recorded Direct Procedures, where we observe a positive correlation with its SHAP values, but only below a specific threshold. For example, for suppliers with less than half of direct procedures, the Pearson correlation is 0.55, while for those with less than 0.9 the correlation is 0.36; and for those with more the 0.9 the correlation turns into negative. This indicates that a higher proportion of direct procedures contributes to a higher predicted probability of fraud, but only for those suppliers whose majority of contracts are non-direct.

\textbf{Supplier Number of Contracts} shows a clear positive relationship with the SHAP values, but only when the proportion of Recorded Direct Procedures is not at its highest level. It is important to note that, although the number of contracts is included as a feature in the model, suppliers within the top 5\% in terms of contract count were undersampled in the training set. Consequently, the SHAP value distribution is not biased by the presence of these highly active suppliers. Furthermore, the plot indicates that suppliers with a large number of contracts generally have higher positive contributions to the predicted probability of fraud; however, this relationship holds primarily for suppliers with a relatively low proportion of Recorded Direct Procedures. This further supports the idea that the model could potentially underestimate risks for suppliers with overwhelming non-competitive tenders.

\textbf{Buyer–Supplier Spending per Active Week} shows a positive correlation with SHAP values, indicating that higher spending concentration over shorter time spans increases the predicted probability of fraud. This variable divides the total spending between a buyer–supplier pair by the number of active weeks in which at least one transaction occurred. When analyzed in interaction with Supplier Coreness, it becomes clear that not all suppliers in the network core exhibit high spending per active week. However, those that combine both medium-high and high coreness and high spending concentration show the largest SHAP values, suggesting that temporal concentration of spending is an important risk factor among core suppliers.

\section*{Conclusion}\label{conclusion}

In this paper we presented a methodology based on Positive-Unlabeled (PU) learning to detect fraud and corruption in Mexican public procurement using labels based on readily available sanctions information of government suppliers. The developed model is comprehensive, incorporating traditional red flags such as the use of direct awards and also relational information embedded in contracting networks, such as supplier centrality.

We demonstrate that suppliers’ sanctions can serve as labels to infer risk factors from, outlining a methodology applicable across a wide array of countries. Neverthless, our approach faces difficult methodological challenges: we only have positive labels, no negative ones; and the observed labels are very rare in the overall dataset. We used two different algorithms to tackle these problems: PU Bagging \cite{mordeletBaggingSVMLearn2014} and Hellinger Distance Stratified Random Forest (HDSRF) \cite{ortegavazquezHellingerDistanceDecision2023} and evaluated their performance based on their average gain and average lift. We observed that the HDSRF model has shown consistently better performance than the other algorithm in the evaluation measures we used and in both transductive and inductive learning. Moreover we also compared ML models to ranking done by the Corruption Risk Index (CRI\cite{fazekas_objective_2016}), which was also surpassed by the HDSRF.

In addition to the performance evaluation, we also unpacked the best-performing model, pointing at surprising variable importances and relationships between fraud risk and risk features. On the whole, network features contribute more to model prediction than traditional red flag indicators, suggesting that the literature should focus more on network methods and features \cite{lyraFraudCorruptionCollusion2022}. Furthermore, established red flags contribute to prediction accuracy most when aggregated to the supplier level, rather than on the contract-level.

We provided an interpretation of our best model by presenting the most important features and examining their relationships with the model’s predictions. We used the SHapley Additive exPlanaitions (SHAP values) \cite{lundbergUnifiedApproachInterpreting2017, lundbergConsistentIndividualizedFeature2019} for interpretation. More specifically for network features, from the suppliers’ perspective, higher contributions to the predicted probability of being a fraudulent contract are associated with (1) being in the core of the bipartite network, (2) being connected to influential buyers and suppliers, (3) have fewer buyer intermediaries to other suppliers, (4) have a large number of suppliers sharing common buyers, and (5) have a large number of shared buyers with neighbors, suggesting coordinated behavior with other suppliers. From the buyers' side, the high positive contributions to the predicted probability tend to involve buyers that (1) are closer to the core of the bipartite network, and (2) are linked to other well-connected buyers; while negative contributions are associated with (3) buyers with fewer intermediaries and (4) with extensive overlap of suppliers with other buyers.

Regarding domain-knowledge features, that is established red flags of corruption, we found a complex, i.e. non-linear, and often weak association with predicted risks of fraud. This aligns with recent literature testing red flags against proven cases, e.g. in Italy by \cite{decarolisCorruptionRedFlags2022}. These findings point at two crucial arguments from prior literature. First, available labels derived from proven cases of corruption, fraud, and related criminal charges are by no means unbiased samples of corrupt behaviours. In some countries, political influence over law enforcement and courts means that some well-connected corrupt actor never get charged while others are charged predominantly for political gain, rather than reflecting the magnitude of their wrongdoing. Yet, in other countries, administrative and legal constraints mean that some types of corruption go unprosecuted. For the latter, we found evidence in Mexico as nearly 75\% of contracts in our dataset follow direct procedure types which are less amenable to audit and investigations due to weaker audit trail \cite{gerardinoDistortionAuditEvidence2024}. For suppliers with substantial non-direct awards (i.e. more than half of contracts), red flags, such as direct award share, have the expected positive relationship with predicted risks.

Second, red flags have been developed to proxy high-level institutionalized forms of corruption (e.g. \cite{fazekas_comprehensive_2016} rather than outright illegal and sanctionable activities. This means that many, yet not all, activities flagged by established red flags are formally legal as high-level political influence assures formal procedural correctness. This argument is supported by literature pointing at the strong association between red flags and personal political connections of suppliers (e.g. \cite{romeroBureaucraticCapacityPolitical2025}), supplier donations (e.g. \cite{titlPoliticalDonationsPublic2021}), and political control over the bureaucracy (e.g. \cite{fazekasAgencyIndependenceCampaign2023}). Based on these two main arguments, it appears beneficial to combine ML-based and traditional red flags and use them in tandem for a more comprehensive measurement. While more research is needed for exploring their alignment as well as their strengths and weaknesses.

In terms of generalization, applying this methodology to other contexts requires several adaptations. First, the model should be trained on contracts within the relevant national context, as the Mexican federal procurement system exhibits specific characteristics that may not apply to other countries and territories. Second, our labeling strategy —which classifies all contracts of a sanctioned company as fraudulent— is only valid under the assumptions that sanctions do not alter company behavior over time and that contracts from sanctioned companies are highly similar across different years. If these conditions do not hold, alternative labeling strategies should be adopted. Third, the strategy of undersampling contracts from large companies in our training test was adopted since the high concentration of contracts in few suppliers impacts the independence assumption of our machine learning models; however, however, this strategy is only appropriate when there's strong similarity between contracts of same supplier, specially for large companies. If there is no such similarity, another strategy should be adapted.

In terms of policy implications, our machine learning approach is replicable and transparent able to detect likely fraudulent contracts based on domain-knowledge and network features, leveraging publicly available information. In practice, it enables the ranking of contracts and the identification of those with the highest predicted probability, providing a concrete and interpretable tool to guide auditing and investigative efforts. We believe this methodology would be useful for Mexican and potentially other authorities in their efforts to investigate fraud and corruption in public procurement.
\newpage

\bibliographystyle{acm}
\bibliography{fraudpractices.bib}

\section*{Author Contributions}
M.M.H., J.K. and M.F. conceived the idea of the paper and wrote the manuscript. MMH collected the data and implemented the methods.

\section*{Data Availability Statement}
Data to reproduce our analysis available upon request.

\section*{Additional Information}
\textbf{Competing Interests:} The authors declare no competing interests

\newpage
\begin{appendices}
\renewcommand{\thesection}{\Alph{section}}
\renewcommand{\appendixname}{Supplementary Section}
\titleformat{\section}
  {\normalfont\Large\bfseries}
  {Supplementary Section~\thesection:}
  {1em}{}

\section{Dataset construction}\label{dataset-construction}

The main dataset used for this paper consist on three sources: contract-level Compranet dataset (c-Compranet), procedure-level Compranet dataset (p-Compranet), and procedure-level 'Open Contracting Data Standard' (p-OCDS) dataset. 

The \emph{c-Compranet dataset} consists on 2,301,278 contracts made by the Mexican federal, state and municipal level governments in a span from 2011 to 2022. The data contains contract-level information of 5,304 Mexican government entities with 259,534 different private suppliers. By law, all Mexican government entities that are using federal-level money are obligated to report the characteristics of the contract to the platform CompraNet \cite{compranet_datos_nodate}. 

The \emph{p-Compranet dataset} contains a synthesis of 1,847,730 different procedures that corresponded to 1,504,429 contracts in c-Compranet (in the Compranet website, this file can be found as 'Expedientes publicados'). It is worth notice that a procedure can have more than one contract, and the matches span over the whole dataset, but they are concentrated in the years 2014-2017, 2019, 2021-2023, as you can see in table \ref{tab:matches-contracts-extendedfiles}. 

Finally the \emph{p-OCDS dataset} also contains the procedures followed by the Mexican authorities to assign a contract, following the OCDS standard \cite{opencontractingpartnershipOpenContractingData}, which has a much more detailed account of the procedures. It contains information of 830,841 different procedures that matched 775,092 contracts (a procedure can correspond to more than one contract). Most of matches however are concentrated in the period of 2018 - 2022, as you can see in table \ref{tab:matches-contracts-ocds}.

\begin{table}[!ht]
\centering
\begin{tabular}{rrrr}
\toprule
Contract Year & N. c-Compranet & N. p-Compranet & Prop. of Matches \\
\midrule
2011 & 124033 & 23 & 0.000185 \\
2012 & 167311 & 241 & 0.001440 \\
2013 & 177946 & 107851 & 0.606088 \\
2014 & 195591 & 119697 & 0.611976 \\
2015 & 220647 & 158860 & 0.719974 \\
2016 & 231246 & 188391 & 0.814678 \\
2017 & 234067 & 211039 & 0.901618 \\
2018 & 194078 & 6736 & 0.034708 \\
2019 & 195128 & 177663 & 0.910495 \\
2020 & 161720 & 6194 & 0.038301 \\
2021 & 206112 & 185776 & 0.901335 \\
2022 & 193399 & 178874 & 0.924896 \\
\bottomrule
\end{tabular}
\caption{Matches between c-Compranet and p-Compranet datasets}
\label{tab:matches-contracts-extendedfiles}
\end{table}
\FloatBarrier
\clearpage
\begin{table}
\centering
\begin{tabular}{rrrr}
\toprule
Contract Year & N. c-Compranet & N. p-OCDS & Prop. of Matches \\
\midrule
2011 & 124033 & NaN & NaN \\
2012 & 167311 & NaN & NaN \\
2013 & 177946 & 1 & 0.000006 \\
2014 & 195591 & NaN & NaN \\
2015 & 220647 & 1 & 0.000005 \\
2016 & 231246 & 21 & 0.000091 \\
2017 & 234067 & 23733 & 0.101394 \\
2018 & 194078 & 144078 & 0.742372 \\
2019 & 195128 & 164568 & 0.843385 \\
2020 & 161720 & 127418 & 0.787893 \\
2021 & 206112 & 151775 & 0.736371 \\
2022 & 193399 & 162659 & 0.841054 \\
\bottomrule
\end{tabular}
\caption{Matches between c-Compranet and p-OCDS datasets}
\label{tab:matches-contracts-ocds}
\end{table}

The \emph{PCS} dataset in our investigation is composed by three different releases: March 2021, April 2023 and February 2024. The March 2021 release was collected by \cite{falcon-cortes_practices_2022}, the rest are from government sources \cite{secretaria_de_la_funcion_publica_proveedores_nodate}.We included all companies in those releases that were proved to have committed some sort of fraud, however, the releases are not cumulative because some of those companies have pursued legal battles to remove their names from the lists.

\newpage
\section{Labelying hypotheses}\label{labelying-hypotheses}

In order to map company-level sanctions to contract-level labels, we examined two hypothesis: (1) the sanctions have a dissuasive effect on the behavior of companies, and (2) contracts of sanctioned companies change significantly over time.

\begin{figure}[!ht]
\centering
\includegraphics[width=1\textwidth]{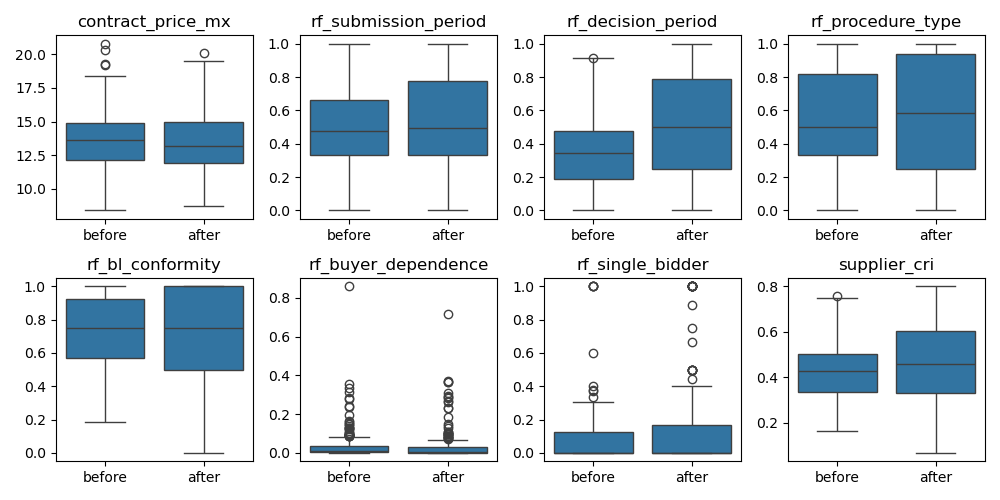}
\caption{One-time offenders' supplier behavior before and after sanction. The figure shows the main red flags of suppliers' contracts and contract price before and after the company was sanctioned. The suppliers correspond to those that had only one sanction in the sanction dataset, therefore, we called one-time offenders. We can observe no appreciable difference in the red flags distribution before and after the sanction. Number }\label{fig:sanctions-onetime}
\end{figure}

\begin{figure}[!ht]
\centering
\includegraphics[width=1\textwidth]{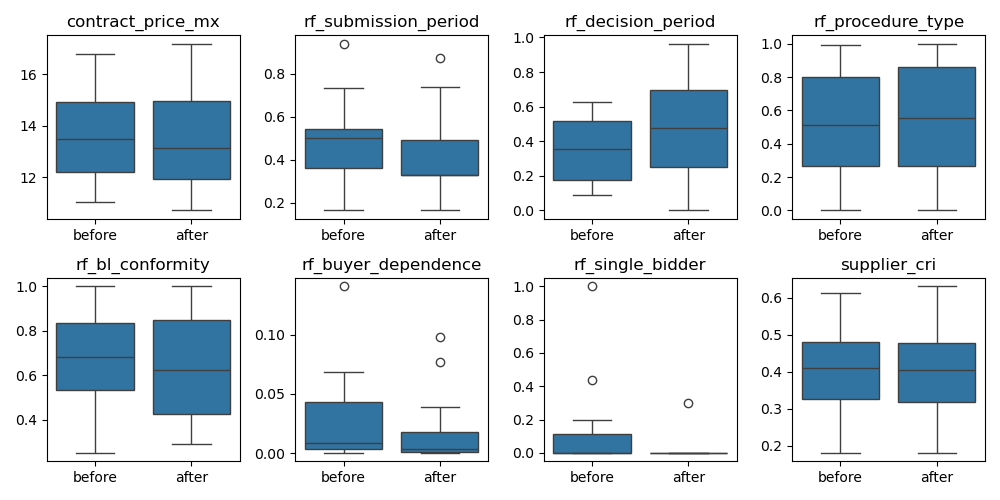}
\caption{Repeated offenders' supplier behavior before and after sanction. The figure shows the main red flags of suppliers' contracts and contract price before and after the company was sanctioned. The suppliers correspond to those that had more than one sanction in different years in the sanction dataset, therefore, we called repeated offenders. We can observe no appreciable difference in the red flags distribution before and after the sanction.}\label{fig:sanctions-repeated}
\end{figure}

In order to check if there's a dissuasive effect on companies behavior, we examined the red flags, CRI \cite{fazekas_objective_2016}, contract price and number of contracts before and after the company was sanctioned. It is worth to mention, that from the 1673 sanctioned companies that were sanctioned, only 283 (16\%) have contracts after the sanction, which indicates that authorities in general avoid to hire past sanctioned companies. From the subset of companies that have contracts after the sanction, we observed no strong differences in the main red flags comparing before and after a sanction for one-time offenders \ref{fig:sanctions-onetime} and repeated offenders \ref{fig:sanctions-repeated}.

\begin{figure}
\centering
\includegraphics[width=1\textwidth]{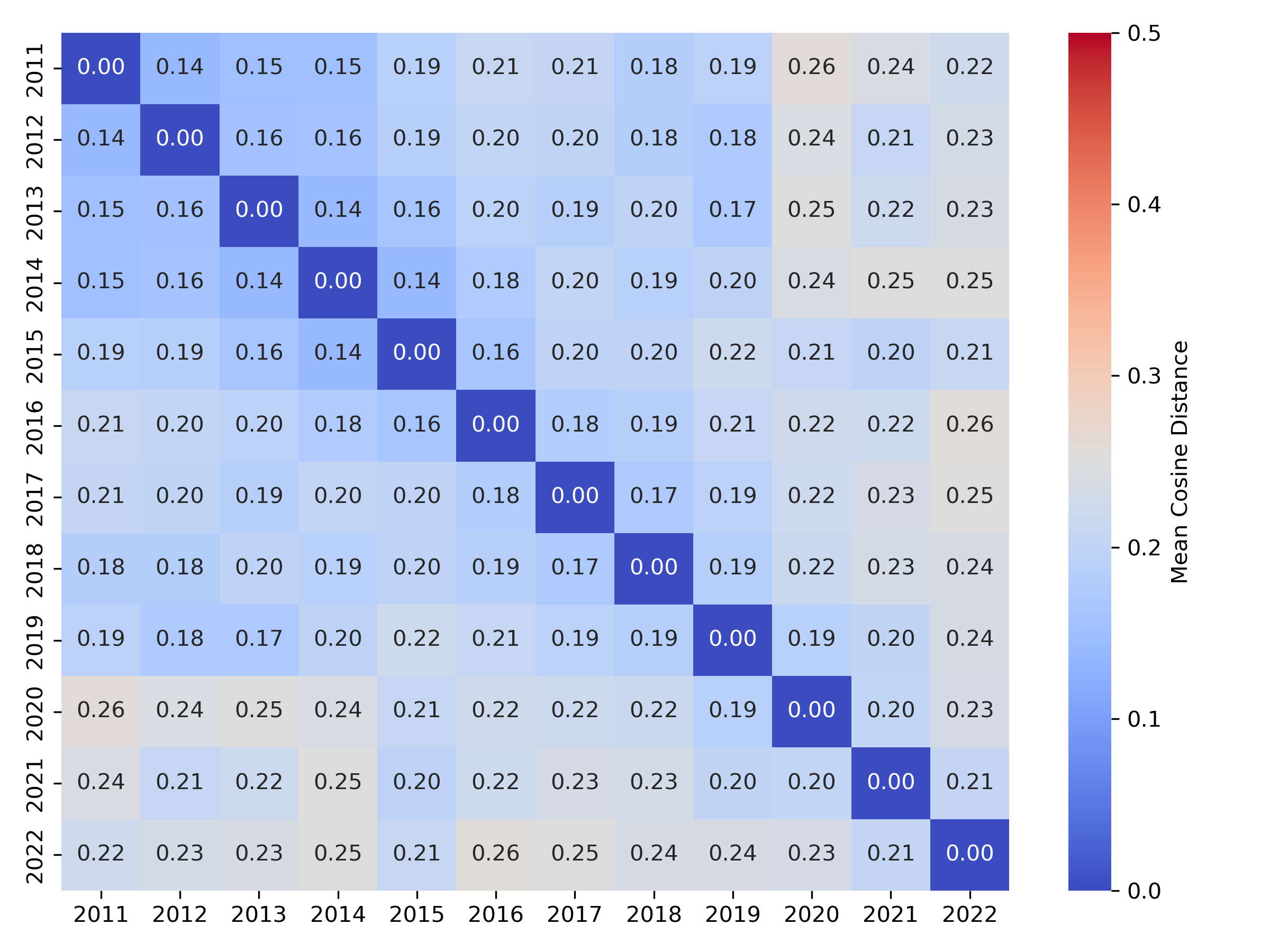}
\caption{Mean cosine distance of contracts of same sanctioned supplier across years. We observe that the mean cosine distance of our contracts is low (below 0.25) in most cases. This means that the sanctioned companies have very similar contracts, even across years.}\label{fig:contract-simmilarity}
\end{figure}

Moreover, for the second hypothesis, we analyzed the observations of sanctioned suppliers in a given year and calculate the average cosine distance with the contracts of the same company in all the other years. This would allow us to see if the contract-level observations are different across years. As you can see in figure \ref{fig:contract-simmilarity}, the mean cosine distance of contracts across the years is very small, indicating that there's few variability in the contracts of same sanctioned supplier across different years.

Our analysis found neither evidence that the sanctions have a dissuasive effect on the behavior of companies nor strong variation among contracts of the same company across time. Following the results of our analysis we decided to consider all contracts of a sanctioned company as examples of a possibly fraudulent contract. This is a strong but reasonable assumption since a company that behaves illegally has incentives to repeatedly do it unless there's are consequences for its behavior.

\FloatBarrier
\clearpage

\clearpage
\newpage
\section{Features Definition}\label{features-definition}

\begin{longtable}{|m{4cm}|m{6cm}|m{2cm}|}
\hline
\textbf{Feature name} & \textbf{Definition} & \textbf{Level} \\ \hline 
\hline
\endfirsthead
\hline
\textbf{Feature name} & \textbf{Definition} & \textbf{Level} \\ \hline 
\hline
\endhead
\hline
\endlastfoot
Buyer Avg. CRI                                    & Sum of contracts' CRI over the total number of contracts of a buyer in a   given year                                                                                                                                                             & Domain Knowledge    \\ \hline
Buyer Number of Contracts                         & Buyer Strength in bipartite network in a given year (i.e. number of   contracts)                                                                                                                                                                  & Domain Knowledge           \\ \hline
Buyer Prop. of Post-Direct Contracts              & Number of post-direct contracts of a buyer over the total number of   contracts of the buyer in a given year                                                                                                                                      & Domain Knowledge    \\ \hline
Buyer Prop. of Real-Direct Contracts              & Number of real-direct contracts of a buyer over the total number of contracts of the buyer in a given year.                                                                                                                                     & Domain Knowledge    \\ \hline
Buyer Prop. of Recorded-Direct Procedures         & Number of direct contracts of a buyer over the total number of contracts   of the buyer in a given year                                                                                                                                           & Domain Knowledge    \\ \hline
Buyer-Supplier Avg. CRI                           & Sum of all contracts' CRI of a buyer-supplier pair over the total number   of contracts of the buyer in a given year                                                                                                                              & Domain Knowledge           \\ \hline
Buyer-supplier active weeks                       & Number of weeks in which a buyer-supplier pair has signed contracts in   a given year                                                                                                                                                           & Domain Knowledge    \\ \hline
Buyer-Supplier Contract Concentration             & Number of contracts of a buyer-supplier pair over the total number of contracts of the same buyer in a given year.                                                                                                                              & Domain Knowledge    \\ \hline
Buyer-Supplier Num. Contracts Odds                & Number of contracts per buyer-supplier pair in a given year over the median of all pairs in a year.                                                                                                                                             & Domain Knowledge    \\ \hline
Buyer-Supplier Num. Contracts per week            & Number of contracts of a buyer-supplier pair in a given week and year                                                                                                                                                                             & Domain Knowledge    \\ \hline
Buyer-Supplier Num. Contracts                     & Edge Weight in bipartite network in a given year (i.e. number of   contracts of a buyer-supplier pair).                                                                                                                                           & Domain Knowledge          \\ \hline
Buyer-Supplier Prop. Recorded-Direct   Procedures & The number of contracts assigned by a direct procedure in a   buyer-supplier pair over the total number of contracts of a buyer in a given   year                                                                                                 & Domain Knowledge           \\ \hline
Buyer-Supplier Spending per Active Week           & Spending of a buyer on the same supplier over active weeks in a given   year                                                                                                                                                                      & Domain Knowledge    \\ \hline
Compliant Submission Period                       & The classification of a submission period as noncompliant, marginal, or   compliant based on the number of days relative to the applicable legal   framework                                                                                      & Domain Knowledge    \\ \hline
Contract Period                                   & Difference in days between the end date and the initial date of a contract.                                                                                                                                                                     & Domain Knowledge    \\ \hline
Contract Price                                    & Amount of money spent under a contract, expressed in Mexican pesos (MXN)                                                                                                                                                                          & Domain Knowledge    \\ \hline
Entropy Buyer Dependence                          & The negative of the product of the buyer dependence with its logarithm   over the logarithm of the total number of suppliers for a buyer in a given   year                                                                                        & Domain Knowledge    \\ \hline
Fragmented Contract Odds                          & For non-open contract, the fragmented contracts odds is the total contract value for a given supplier-buyer-week divided by the median open   contract value for that procedure type and year.                                                  & Domain Knowledge    \\ \hline
Government level                                  & Buyer level of government that signed the contract                                                                                                                                                                                                & Domain Knowledge    \\ \hline
Legal Framework                                   & Legal framework in which the contract was made                                                                                                                                                                                                    & Domain Knowledge    \\ \hline
Legal Fundament                                   & Law that authorities fundament their decision for not using an open   procedure                                                                                                                                                                   & Domain Knowledge    \\ \hline
MAD                                               & MAD (Mean Absolute Deviation) is the average of the absolute differences   between the observed digit frequencies and the expected frequencies of   contract price for a buyer's contracts in a given year according to Benford’s   distribution. & Domain Knowledge    \\ \hline
Neighborhood Avg. CRI                             & Sum of the average CRI of the neighbours of a buyer-supplier pair over   the total number of neighbours in a buyer-supplier bipartite network in a   given year                                                                                   & Domain Knowledge      \\ \hline
Neighborhood Prop. Recorded-Direct   Procedures   & Sum of the proportion of direct contracts of the neighbours of a   buyer-supplier pair over the total number of neighbours in a buyer-supplier   bipartite network in a given year                                                                & Domain Knowledge           \\ \hline
Not in National Registry                          & Abscense of supplier in the Mexican National Registry of Suppliers (RUPC)                                                                                                                                                                         & Domain Knowledge    \\ \hline
Procedure venue                                   & Venue that was used to make the contract                                                                                                                                                                                                          & Domain Knowledge    \\ \hline
R.F. Benford Law Conformity                       & Categorization of the Mean Absolute Deviation (MAD) between the distribution of buyers’ contract values and the Benford’s Law distribution, computed for buyers whose number of contracts exceeds the 0.75 quantile of all buyers in a given year e.g. buyers with more than  42–62 contracts depending on the year.                                                                                                                                                                                                                 & Domain Knowledge    \\ \hline
R.F. Buyer Dependence                             & Contract value granted to a supplier in a year divided by the total amount a buyer has spent in the same year                                                                                                                                                                                                                      & Domain Knowledge    \\ \hline
R.F. Decision Period                              & Number of days between submission deadline and contract award date                                                                                                                                                                                                                        & Domain Knowledge    \\ \hline
R.F. Procedure Type                               & Type of legal procedure followed by the authorities to assign a contract                                                                                                                                                                                                                         & Domain Knowledge    \\ \hline
R.F. Single Bidder                                & Procurement procedure with only one bidder                                                                                                                                                                                                                          & Domain Knowledge    \\ \hline
R.F. Submission Period                            & Number of days between submission deadline and call for tender publication                                                                                                                                                                                                                      & Domain Knowledge    \\ \hline
Supplier Avg. CRI                                 & Sum of all supplier’s contract CRIs divided by their number of contracts   in a given year                                                                                                                                                        & Domain Knowledge    \\ \hline
Supplier Number of Contracts                      & Supplier Strength in bipartite network in a given year (i.e. number of   contracts)                                                                                                                                                               & Domain Knowledge         \\ \hline
Supplier Prop. of Post-Direct Procedures          & Number of post-direct contracts of a supplier over the total number of   contracts of the buyer in a given year                                                                                                                                   & Domain Knowledge    \\ \hline
Supplier Prop. of Real Direct Procedures          & Number of real-direct contracts of a supplier over the total number of   contracts of the buyer in a given year.                                                                                                                                  & Domain Knowledge    \\ \hline
Supplier Prop. of Recorded Direct   Procedures    & Number of direct contracts of a supplier over the total number of   contracts of the buyer in a given year                                                                                                                                        & Domain Knowledge    \\ \hline
Supplier Size                                     & Self-adscription of company size                                                                                                                                                                                                                  & Domain Knowledge    \\ \hline
Supply Type                                       & Type of supply agreed in the contract                                                                                                                                                                                                             & Domain Knowledge    \\ \hline
Tender Period                                     & Difference in days between the award date and the advertisement date of   the contract                                                                                                                                                            & Domain Knowledge    \\ \hline
Buyer Coreness (avg. degree)                      & The highest k-shell (or core) a buyer node belongs to obtained from the   average-degree k-shell decomposition of the bipartite network.                                                                                                          & Bipartite           \\ \hline
Buyer Coreness (strength)                         & The highest k-shell (or core) a buyer node belongs to obtained from the   strength k-shell decomposition of the bipartite network.                                                                                                                & Bipartite           \\ \hline
Buyer Coreness (weighted degree)                  & The highest k-shell (or core) a buyer node belongs to obtained from the   weighted-degree k-shell decomposition of the bipartite network.                                                                                                         & Bipartite           \\ \hline
Competitive Clustering (unweighted)               & The number of cycles over the number of paths of length three for a buyer   in the buyer-supplier bipartite network of a given year                                                                                                               & Bipartite           \\ \hline
Competitive Clustering (weighted,   bounded)      & The Weighted Competitive Clustering (Unbounded) over the number of paths   of length three of a buyer in the bipartite network in a given year                                                                                                    & Bipartite           \\ \hline
Competitive Clustering (weighted,   unbounded)    & The product of the competitive clustering with the sum of each geometric   mean of the scaled edge weights of each cycle of a buyer in a buyer-supplier   bipartite network in a given year.                                                      & Bipartite           \\ \hline
Edge Betweenness Centrality                       & Edge Betweenness Centrality in the bipartite network in a given year   (normalized)                                                                                                                                                               & Bipartite           \\ \hline
Supplier Coreness (avg. degree)                   & The highest k-shell (or core) a supplier node belongs to obtained from   the average-degree k-shell decomposition of the bipartite network.                                                                                                       & Bipartite           \\ \hline
Supplier Coreness (strength)                      & The highest k-shell (or core) a supplier node belongs to obtained from   the stength k-shell decomposition of the bipartite network.                                                                                                              & Bipartite           \\ \hline
Supplier Coreness (weighted degree)               & The highest k-shell (or core) a supplier node belongs to obtained from   the weighted-degree k-shell decomposition of the bipartite network.                                                                                                      & Bipartite           \\ \hline

Buyer Betweenness Centrality                      & Node Betweenness Centrality in buyer projection of the bipartite network   in a given year (normalized)                                                                                                                                           & Buyer Projection    \\ \hline
Buyer Closeness Centrality                        & Closeness Centrality in the buyer projection of the bipartite network in   a given year (normalized)                                                                                                                                              & Buyer Projection    \\ \hline
Buyer Degree Centrality                           & Degree Centrality in the buyer projection of the bipartite network in a   given year (normalized)                                                                                                                                                 & Buyer Projection    \\ \hline
Buyer Eigenvector Centrality                      & Eigenvector Centrality in the buyer projection of the bipartite network   in a given year (normalized)                                                                                                                                            & Buyer Projection    \\ \hline
Buyer Strength                                    & Buyer Strength in buyer projection of bipartite network in a given year                                                                                                                                                                           & Buyer Projection    \\ \hline
Supplier Betweenness Centrality                   & Node Betweenness Centrality in supplier projection of the bipartite   network in a given year (normalized)                                                                                                                                        & Supplier Projection \\ \hline
Supplier Closeness Centrality                     & Closeness Centrality in the supplier projection of the bipartite network   in a given year (normalized)                                                                                                                                           & Supplier Projection \\ \hline
Supplier Degree Centrality                        & Degree Centrality in the supplier projection of the bipartite network in   a given year (normalized)                                                                                                                                              & Supplier Projection \\ \hline
Supplier Eigenvector Centrality                   & Eigenvector Centrality in the supplier projection of the bipartite   network in a given year (normalized)                                                                                                                                         & Supplier Projection \\ \hline
Supplier Strength                                 & Supplier Strength in supplier projection of bipartite network in a given   year                                                                                                                                                                   & Supplier Projection \\ \hline
\caption{Features' Definition. The table shows the features' definition and level of analysis.}\label{tab:feature-definition} 
\end{longtable}

\newpage
\section{Data Frame Summary}\label{dfsummary}
{
\scriptsize
\begin{longtable}{|@{}p{0.3cm}|p{1.8cm}|p{3.7cm}|p{1.8cm}|p{3cm}|p{1cm}|}
\hline
\textbf{No} & \textbf{Variable} & \textbf{Stats / Values} & \textbf{Freqs (\%)} & \textbf{Graph} & \textbf{M.}  \\ \hline
\begin{itemize}[itemsep=4pt, left=0pt, label={}]
\item 1
\end{itemize}
&
\begin{itemize}[itemsep=4pt, left=0pt, label={}]
\item Buyer-supplier active weeks
\item \textnormal{[integer]}
\end{itemize}
&
\begin{itemize}[noitemsep, left=0pt, label={}]
\item Mean (sd):
\item 8.9 (13.9)
\item min \textless med \textless max:
\item 1 \textless 2 \textless 52
\item IQR (CV):
\item 7 (1.6)
\end{itemize}
&
\begin{itemize}[noitemsep, left=0pt, label={}]
\item 52
\item distinct
\item values
\end{itemize}
&
\raisebox{-0.9\totalheight}{\includegraphics[width=3cm]{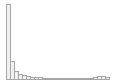}}
&
\begin{itemize}[noitemsep, left=-0.1pt, label={}]
\item 0.0\%
\end{itemize}
\\
\hline
\begin{itemize}[itemsep=4pt, left=0pt, label={}]
\item 2
\end{itemize}
&
\begin{itemize}[itemsep=4pt, left=0pt, label={}]
\item Buyer-Supplier Avg. CRI
\item \textnormal{[numeric]}
\end{itemize}
&
\begin{itemize}[noitemsep, left=0pt, label={}]
\item Mean (sd):
\item 0.6 (0.2)
\item min \textless med \textless max:
\item 0 \textless 0.6 \textless 1
\item IQR (CV):
\item 0.3 (0.3)
\end{itemize}
&
\begin{itemize}[noitemsep, left=0pt, label={}]
\item 971298
\item distinct
\item values
\end{itemize}
&
\raisebox{-0.9\totalheight}{\includegraphics[width=3cm]{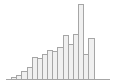}}
&
\begin{itemize}[noitemsep, left=-0.1pt, label={}]
\item 0.0\%
\end{itemize}
\\
\hline
\begin{itemize}[itemsep=4pt, left=0pt, label={}]
\item 3
\end{itemize}
&
\begin{itemize}[itemsep=4pt, left=0pt, label={}]
\item Buyer-Supplier Contract Concentration
\item \textnormal{[numeric]}
\end{itemize}
&
\begin{itemize}[noitemsep, left=0pt, label={}]
\item Mean (sd):
\item 0 (0.1)
\item min \textless med \textless max:
\item 0 \textless 0 \textless 1
\item IQR (CV):
\item 0 (2.4)
\end{itemize}
&
\begin{itemize}[noitemsep, left=0pt, label={}]
\item 14777
\item distinct
\item values
\end{itemize}
&
\raisebox{-0.9\totalheight}{\includegraphics[width=3cm]{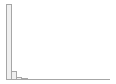}}
&
\begin{itemize}[noitemsep, left=-0.1pt, label={}]
\item 0.0\%
\end{itemize}
\\
\hline
\begin{itemize}[itemsep=4pt, left=0pt, label={}]
\item 4
\end{itemize}
&
\begin{itemize}[itemsep=4pt, left=0pt, label={}]
\item Buyer-Supplier Num. Contracts
\item \textnormal{[integer]}
\end{itemize}
&
\begin{itemize}[noitemsep, left=0pt, label={}]
\item Mean (sd):
\item 74.2 (260)
\item min \textless med \textless max:
\item 1 \textless 3 \textless 2641
\item IQR (CV):
\item 13 (3.5)
\end{itemize}
&
\begin{itemize}[noitemsep, left=0pt, label={}]
\item 460
\item distinct
\item values
\end{itemize}
&
\raisebox{-0.9\totalheight}{\includegraphics[width=3cm]{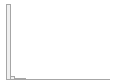}}
&
\begin{itemize}[noitemsep, left=-0.1pt, label={}]
\item 0.0\%
\end{itemize}
\\
\hline
\begin{itemize}[itemsep=4pt, left=0pt, label={}]
\item 5
\end{itemize}
&
\begin{itemize}[itemsep=4pt, left=0pt, label={}]
\item Buyer-Supplier Num. Contracts Odds
\item \textnormal{[numeric]}
\end{itemize}
&
\begin{itemize}[noitemsep, left=0pt, label={}]
\item Mean (sd):
\item 25.5 (91)
\item min \textless med \textless max:
\item 0.2 \textless 1 \textless 880.3
\item IQR (CV):
\item 4 (3.6)
\end{itemize}
&
\begin{itemize}[noitemsep, left=0pt, label={}]
\item 817
\item distinct
\item values
\end{itemize}
&
\raisebox{-0.9\totalheight}{\includegraphics[width=3cm]{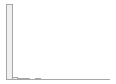}}
&
\begin{itemize}[noitemsep, left=-0.1pt, label={}]
\item 0.0\%
\end{itemize}
\\
\hline
\begin{itemize}[itemsep=4pt, left=0pt, label={}]
\item 6
\end{itemize}
&
\begin{itemize}[itemsep=4pt, left=0pt, label={}]
\item Buyer-Supplier Num. Contracts per week
\item \textnormal{[integer]}
\end{itemize}
&
\begin{itemize}[noitemsep, left=0pt, label={}]
\item Mean (sd):
\item 5.1 (28.4)
\item min \textless med \textless max:
\item 1 \textless 1 \textless 706
\item IQR (CV):
\item 1 (5.6)
\end{itemize}
&
\begin{itemize}[noitemsep, left=0pt, label={}]
\item 129
\item distinct
\item values
\end{itemize}
&
\raisebox{-0.9\totalheight}{\includegraphics[width=3cm]{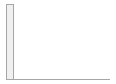}}
&
\begin{itemize}[noitemsep, left=-0.1pt, label={}]
\item 0.0\%
\end{itemize}
\\
\hline
\begin{itemize}[itemsep=4pt, left=0pt, label={}]
\item 7
\end{itemize}
&
\begin{itemize}[itemsep=4pt, left=0pt, label={}]
\item Buyer-Supplier Prop. Recorded-Direct Procedures
\item \textnormal{[numeric]}
\end{itemize}
&
\begin{itemize}[noitemsep, left=0pt, label={}]
\item Mean (sd):
\item 0.7 (0.4)
\item min \textless med \textless max:
\item 0 \textless 1 \textless 1
\item IQR (CV):
\item 0.5 (0.5)
\end{itemize}
&
\begin{itemize}[noitemsep, left=0pt, label={}]
\item 696
\item distinct
\item values
\end{itemize}
&
\raisebox{-0.9\totalheight}{\includegraphics[width=3cm]{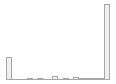}}
&
\begin{itemize}[noitemsep, left=-0.1pt, label={}]
\item 0.0\%
\end{itemize}
\\
\hline
\begin{itemize}[itemsep=4pt, left=0pt, label={}]
\item 8
\end{itemize}
&
\begin{itemize}[itemsep=4pt, left=0pt, label={}]
\item Buyer-Supplier Spending per Active Week
\item \textnormal{[numeric]}
\end{itemize}
&
\begin{itemize}[noitemsep, left=0pt, label={}]
\item Mean (sd):
\item \(3.8 \times 10^6\) (\(8.6 \times 10^7\))
\item min \textless med \textless max:
\item 0 \textless \(2.8 \times 10^5\) \textless \(8.5 \times 10^{10}\)
\item IQR (CV):
\item \(1.2 \times 10^6\) (22.8)
\end{itemize}
&
\begin{itemize}[noitemsep, left=0pt, label={}]
\item 703876
\item distinct
\item values
\end{itemize}
&
\raisebox{-0.9\totalheight}{\includegraphics[width=3cm]{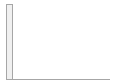}}
&
\begin{itemize}[noitemsep, left=-0.1pt, label={}]
\item 0.0\%
\end{itemize}
\\
\hline
\begin{itemize}[itemsep=4pt, left=0pt, label={}]
\item 9
\end{itemize}
&
\begin{itemize}[itemsep=4pt, left=0pt, label={}]
\item Buyer Avg. CRI
\item \textnormal{[numeric]}
\end{itemize}
&
\begin{itemize}[noitemsep, left=0pt, label={}]
\item Mean (sd):
\item 0.6 (0.2)
\item min \textless med \textless max:
\item 0.1 \textless 0.6 \textless 0.9
\item IQR (CV):
\item 0.2 (0.3)
\end{itemize}
&
\begin{itemize}[noitemsep, left=0pt, label={}]
\item 20751
\item distinct
\item values
\end{itemize}
&
\raisebox{-0.9\totalheight}{\includegraphics[width=3cm]{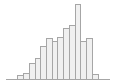}}
&
\begin{itemize}[noitemsep, left=-0.1pt, label={}]
\item 0.0\%
\end{itemize}
\\
\hline
\begin{itemize}[itemsep=4pt, left=0pt, label={}]
\item 10
\end{itemize}
&
\begin{itemize}[itemsep=4pt, left=0pt, label={}]
\item Buyer Betweenness Centrality
\item \textnormal{[numeric]}
\end{itemize}
&
\begin{itemize}[noitemsep, left=0pt, label={}]
\item Mean (sd):
\item 0 (0)
\item min \textless med \textless max:
\item 0 \textless 0 \textless 0.1
\item IQR (CV):
\item 0 (2.3)
\end{itemize}
&
\begin{itemize}[noitemsep, left=0pt, label={}]
\item 19457
\item distinct
\item values
\end{itemize}
&
\raisebox{-0.9\totalheight}{\includegraphics[width=3cm]{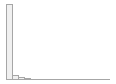}}
&
\begin{itemize}[noitemsep, left=-0.1pt, label={}]
\item 0.0\%
\end{itemize}
\\
\hline
\begin{itemize}[itemsep=4pt, left=0pt, label={}]
\item 11
\end{itemize}
&
\begin{itemize}[itemsep=4pt, left=0pt, label={}]
\item Buyer Closeness Centrality
\item \textnormal{[numeric]}
\end{itemize}
&
\begin{itemize}[noitemsep, left=0pt, label={}]
\item Mean (sd):
\item 0.5 (0.1)
\item min \textless med \textless max:
\item 0 \textless 0.5 \textless 0.6
\item IQR (CV):
\item 0.1 (0.1)
\end{itemize}
&
\begin{itemize}[noitemsep, left=0pt, label={}]
\item 17671
\item distinct
\item values
\end{itemize}
&
\raisebox{-0.9\totalheight}{\includegraphics[width=3cm]{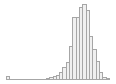}}
&
\begin{itemize}[noitemsep, left=-0.1pt, label={}]
\item 0.0\%
\end{itemize}
\\
\hline
\begin{itemize}[itemsep=4pt, left=0pt, label={}]
\item 12
\end{itemize}
&
\begin{itemize}[itemsep=4pt, left=0pt, label={}]
\item Buyer Coreness (avg. degree)
\item \textnormal{[numeric]}
\end{itemize}
&
\begin{itemize}[noitemsep, left=0pt, label={}]
\item Mean (sd):
\item 0.2 (0.3)
\item min \textless med \textless max:
\item 0 \textless 0 \textless 1
\item IQR (CV):
\item 0.2 (1.5)
\end{itemize}
&
\begin{itemize}[noitemsep, left=0pt, label={}]
\item 510
\item distinct
\item values
\end{itemize}
&
\raisebox{-0.9\totalheight}{\includegraphics[width=3cm]{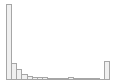}}
&
\begin{itemize}[noitemsep, left=-0.1pt, label={}]
\item 0.0\%
\end{itemize}
\\
\hline
\begin{itemize}[itemsep=4pt, left=0pt, label={}]
\item 13
\end{itemize}
&
\begin{itemize}[itemsep=4pt, left=0pt, label={}]
\item Buyer Coreness (strength)
\item \textnormal{[numeric]}
\end{itemize}
&
\begin{itemize}[noitemsep, left=0pt, label={}]
\item Mean (sd):
\item 0.4 (0.4)
\item min \textless med \textless max:
\item 0 \textless 0.2 \textless 1
\item IQR (CV):
\item 0.7 (0.9)
\end{itemize}
&
\begin{itemize}[noitemsep, left=0pt, label={}]
\item 1008
\item distinct
\item values
\end{itemize}
&
\raisebox{-0.9\totalheight}{\includegraphics[width=3cm]{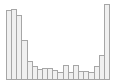}}
&
\begin{itemize}[noitemsep, left=-0.1pt, label={}]
\item 0.0\%
\end{itemize}
\\
\hline
\begin{itemize}[itemsep=4pt, left=0pt, label={}]
\item 14
\end{itemize}
&
\begin{itemize}[itemsep=4pt, left=0pt, label={}]
\item Buyer Coreness (weighted degree)
\item \textnormal{[numeric]}
\end{itemize}
&
\begin{itemize}[noitemsep, left=0pt, label={}]
\item Mean (sd):
\item 0.5 (0.3)
\item min \textless med \textless max:
\item 0 \textless 0.4 \textless 1
\item IQR (CV):
\item 0.7 (0.7)
\end{itemize}
&
\begin{itemize}[noitemsep, left=0pt, label={}]
\item 487
\item distinct
\item values
\end{itemize}
&
\raisebox{-0.9\totalheight}{\includegraphics[width=3cm]{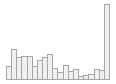}}
&
\begin{itemize}[noitemsep, left=-0.1pt, label={}]
\item 0.0\%
\end{itemize}
\\
\hline
\begin{itemize}[itemsep=4pt, left=0pt, label={}]
\item 15
\end{itemize}
&
\begin{itemize}[itemsep=4pt, left=0pt, label={}]
\item Buyer Degree Centrality
\item \textnormal{[numeric]}
\end{itemize}
&
\begin{itemize}[noitemsep, left=0pt, label={}]
\item Mean (sd):
\item 0.1 (0.1)
\item min \textless med \textless max:
\item 0 \textless 0.1 \textless 0.4
\item IQR (CV):
\item 0.1 (0.7)
\end{itemize}
&
\begin{itemize}[noitemsep, left=0pt, label={}]
\item 5069
\item distinct
\item values
\end{itemize}
&
\raisebox{-0.9\totalheight}{\includegraphics[width=3cm]{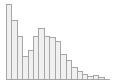}}
&
\begin{itemize}[noitemsep, left=-0.1pt, label={}]
\item 0.0\%
\end{itemize}
\\
\hline
\begin{itemize}[itemsep=4pt, left=0pt, label={}]
\item 16
\end{itemize}
&
\begin{itemize}[itemsep=4pt, left=0pt, label={}]
\item Buyer Eigenvector Centrality
\item \textnormal{[numeric]}
\end{itemize}
&
\begin{itemize}[noitemsep, left=0pt, label={}]
\item Mean (sd):
\item 0 (0.1)
\item min \textless med \textless max:
\item 0 \textless 0 \textless 0.2
\item IQR (CV):
\item 0 (1.5)
\end{itemize}
&
\begin{itemize}[noitemsep, left=0pt, label={}]
\item 24266
\item distinct
\item values
\end{itemize}
&
\raisebox{-0.9\totalheight}{\includegraphics[width=3cm]{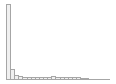}}
&
\begin{itemize}[noitemsep, left=-0.1pt, label={}]
\item 0.0\%
\end{itemize}
\\
\hline
\begin{itemize}[itemsep=4pt, left=0pt, label={}]
\item 17
\end{itemize}
&
\begin{itemize}[itemsep=4pt, left=0pt, label={}]
\item Buyer Number of Contracts
\item \textnormal{[integer]}
\end{itemize}
&
\begin{itemize}[noitemsep, left=0pt, label={}]
\item Mean (sd):
\item 3524.8 (7803.6)
\item min \textless med \textless max:
\item 1 \textless 471 \textless 31135
\item IQR (CV):
\item 1445 (2.2)
\end{itemize}
&
\begin{itemize}[noitemsep, left=0pt, label={}]
\item 1058
\item distinct
\item values
\end{itemize}
&
\raisebox{-0.9\totalheight}{\includegraphics[width=3cm]{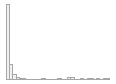}}
&
\begin{itemize}[noitemsep, left=-0.1pt, label={}]
\item 0.0\%
\end{itemize}
\\
\hline
\begin{itemize}[itemsep=4pt, left=0pt, label={}]
\item 18
\end{itemize}
&
\begin{itemize}[itemsep=4pt, left=0pt, label={}]
\item Buyer Prop. of Post-Direct Contracts
\item \textnormal{[numeric]}
\end{itemize}
&
\begin{itemize}[noitemsep, left=0pt, label={}]
\item Mean (sd):
\item 0.5 (0.4)
\item min \textless med \textless max:
\item 0 \textless 0.5 \textless 1
\item IQR (CV):
\item 0.9 (0.8)
\end{itemize}
&
\begin{itemize}[noitemsep, left=0pt, label={}]
\item 4227
\item distinct
\item values
\end{itemize}
&
\raisebox{-0.9\totalheight}{\includegraphics[width=3cm]{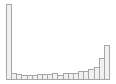}}
&
\begin{itemize}[noitemsep, left=-0.1pt, label={}]
\item 0.0\%
\end{itemize}
\\
\hline
\begin{itemize}[itemsep=4pt, left=0pt, label={}]
\item 19
\end{itemize}
&
\begin{itemize}[itemsep=4pt, left=0pt, label={}]
\item Buyer Prop. of Real-Direct Contracts
\item \textnormal{[numeric]}
\end{itemize}
&
\begin{itemize}[noitemsep, left=0pt, label={}]
\item Mean (sd):
\item 0.3 (0.4)
\item min \textless med \textless max:
\item 0 \textless 0 \textless 1
\item IQR (CV):
\item 0.6 (1.4)
\end{itemize}
&
\begin{itemize}[noitemsep, left=0pt, label={}]
\item 2956
\item distinct
\item values
\end{itemize}
&
\raisebox{-0.9\totalheight}{\includegraphics[width=3cm]{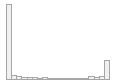}}
&
\begin{itemize}[noitemsep, left=-0.1pt, label={}]
\item 0.0\%
\end{itemize}
\\
\hline
\begin{itemize}[itemsep=4pt, left=0pt, label={}]
\item 20
\end{itemize}
&
\begin{itemize}[itemsep=4pt, left=0pt, label={}]
\item Buyer Prop. of Recorded-Direct Procedures
\item \textnormal{[numeric]}
\end{itemize}
&
\begin{itemize}[noitemsep, left=0pt, label={}]
\item Mean (sd):
\item 0.8 (0.3)
\item min \textless med \textless max:
\item 0 \textless 0.9 \textless 1
\item IQR (CV):
\item 0.3 (0.4)
\end{itemize}
&
\begin{itemize}[noitemsep, left=0pt, label={}]
\item 4561
\item distinct
\item values
\end{itemize}
&
\raisebox{-0.9\totalheight}{\includegraphics[width=3cm]{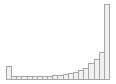}}
&
\begin{itemize}[noitemsep, left=-0.1pt, label={}]
\item 0.0\%
\end{itemize}
\\
\hline
\begin{itemize}[itemsep=4pt, left=0pt, label={}]
\item 21
\end{itemize}
&
\begin{itemize}[itemsep=4pt, left=0pt, label={}]
\item Buyer Strength
\item \textnormal{[numeric]}
\end{itemize}
&
\begin{itemize}[noitemsep, left=0pt, label={}]
\item Mean (sd):
\item 0 (0)
\item min \textless med \textless max:
\item 0 \textless 0 \textless 0
\item IQR (CV):
\item 0 (1.6)
\end{itemize}
&
\begin{itemize}[noitemsep, left=0pt, label={}]
\item 7642
\item distinct
\item values
\end{itemize}
&
\raisebox{-0.9\totalheight}{\includegraphics[width=3cm]{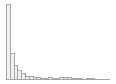}}
&
\begin{itemize}[noitemsep, left=-0.1pt, label={}]
\item 0.0\%
\end{itemize}
\\
\hline
\begin{itemize}[itemsep=4pt, left=0pt, label={}]
\item 22
\end{itemize}
&
\begin{itemize}[itemsep=4pt, left=0pt, label={}]
\item Competitive Clustering (unweighted)
\item \textnormal{[numeric]}
\end{itemize}
&
\begin{itemize}[noitemsep, left=0pt, label={}]
\item Mean (sd):
\item 0.1 (0.1)
\item min \textless med \textless max:
\item 0 \textless 0 \textless 1
\item IQR (CV):
\item 0.2 (1.2)
\end{itemize}
&
\begin{itemize}[noitemsep, left=0pt, label={}]
\item 14934
\item distinct
\item values
\end{itemize}
&
\raisebox{-0.9\totalheight}{\includegraphics[width=3cm]{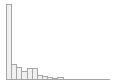}}
&
\begin{itemize}[noitemsep, left=-0.1pt, label={}]
\item 0.0\%
\end{itemize}
\\
\hline
\begin{itemize}[itemsep=4pt, left=0pt, label={}]
\item 23
\end{itemize}
&
\begin{itemize}[itemsep=4pt, left=0pt, label={}]
\item Competitive Clustering (weighted, bounded)
\item \textnormal{[numeric]}
\end{itemize}
&
\begin{itemize}[noitemsep, left=0pt, label={}]
\item Mean (sd):
\item 0 (0)
\item min \textless med \textless max:
\item 0 \textless 0 \textless 0
\item IQR (CV):
\item 0 (1.6)
\end{itemize}
&
\begin{itemize}[noitemsep, left=0pt, label={}]
\item 16634
\item distinct
\item values
\end{itemize}
&
\raisebox{-0.9\totalheight}{\includegraphics[width=3cm]{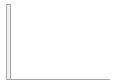}}
&
\begin{itemize}[noitemsep, left=-0.1pt, label={}]
\item 0.0\%
\end{itemize}
\\
\hline
\begin{itemize}[itemsep=4pt, left=0pt, label={}]
\item 24
\end{itemize}
&
\begin{itemize}[itemsep=4pt, left=0pt, label={}]
\item Competitive Clustering (weighted, unbounded)
\item \textnormal{[numeric]}
\end{itemize}
&
\begin{itemize}[noitemsep, left=0pt, label={}]
\item Mean (sd):
\item 0 (0.1)
\item min \textless med \textless max:
\item 0 \textless 0 \textless 0.8
\item IQR (CV):
\item 0 (2.6)
\end{itemize}
&
\begin{itemize}[noitemsep, left=0pt, label={}]
\item 16634
\item distinct
\item values
\end{itemize}
&
\raisebox{-0.9\totalheight}{\includegraphics[width=3cm]{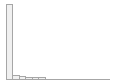}}
&
\begin{itemize}[noitemsep, left=-0.1pt, label={}]
\item 0.0\%
\end{itemize}
\\
\hline
\begin{itemize}[itemsep=4pt, left=0pt, label={}]
\item 25
\end{itemize}
&
\begin{itemize}[itemsep=4pt, left=0pt, label={}]
\item Compliant Submission Period
\item \textnormal{[character]}
\end{itemize}
&
\begin{itemize}[itemsep=4pt, left=0pt, label={}]
\item 1. compliant
\item 2. marginal
\item 3. noncompliant
\end{itemize}
&
\begin{itemize}[itemsep=4pt, left=0pt, label={}]
\item 16.9\%
\item 7.3\%
\item 75.8\%
\end{itemize}
&
\raisebox{-0.9\totalheight}{\includegraphics[height=1.5cm]{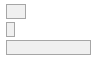}}
&
\begin{itemize}[noitemsep, left=0pt, label={}]
\item 42.3\%
\end{itemize}
\\
\hline
\begin{itemize}[itemsep=4pt, left=0pt, label={}]
\item 26
\end{itemize}
&
\begin{itemize}[itemsep=4pt, left=0pt, label={}]
\item Contract Period
\item \textnormal{[numeric]}
\end{itemize}
&
\begin{itemize}[noitemsep, left=0pt, label={}]
\item Mean (sd):
\item 111.3 (167.9)
\item min \textless med \textless max:
\item -63 \textless 42 \textless 1679
\item IQR (CV):
\item 148 (1.5)
\end{itemize}
&
\begin{itemize}[noitemsep, left=0pt, label={}]
\item 1719
\item distinct
\item values
\end{itemize}
&
\raisebox{-0.9\totalheight}{\includegraphics[width=3cm]{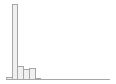}}
&
\begin{itemize}[noitemsep, left=-0.1pt, label={}]
\item 0.0\%
\end{itemize}
\\
\hline
\begin{itemize}[itemsep=4pt, left=0pt, label={}]
\item 27
\end{itemize}
&
\begin{itemize}[itemsep=4pt, left=0pt, label={}]
\item Contract Price
\item \textnormal{[numeric]}
\end{itemize}
&
\begin{itemize}[noitemsep, left=0pt, label={}]
\item Mean (sd):
\item \(2.3 \times 10^6\) (\(8.3 \times 10^7\))
\item min \textless med \textless max:
\item 0 \textless \(1.2 \times 10^5\) \textless \(8.5 \times 10^{10}\)
\item IQR (CV):
\item \(4.3 \times 10^5\) (35.6)
\end{itemize}
&
\begin{itemize}[noitemsep, left=0pt, label={}]
\item 1229348
\item distinct
\item values
\end{itemize}
&
\raisebox{-0.9\totalheight}{\includegraphics[width=3cm]{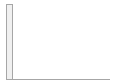}}
&
\begin{itemize}[noitemsep, left=-0.1pt, label={}]
\item 0.0\%
\end{itemize}
\\
\hline
\begin{itemize}[itemsep=4pt, left=0pt, label={}]
\item 28
\end{itemize}
&
\begin{itemize}[itemsep=4pt, left=0pt, label={}]
\item Edge Betweenness Centrality
\item \textnormal{[numeric]}
\end{itemize}
&
\begin{itemize}[noitemsep, left=0pt, label={}]
\item Mean (sd):
\item \(9.8 \times 10^{13}\) (\(3.7 \times 10^{14}\))
\item min \textless med \textless max:
\item 0 \textless \(3.4 \times 10^{13}\) \textless \(6.2 \times 10^{16}\)
\item IQR (CV):
\item \(1.2 \times 10^{14}\) (3.8)
\end{itemize}
&
\begin{itemize}[noitemsep, left=0pt, label={}]
\item 472810
\item distinct
\item values
\end{itemize}
&
\raisebox{-0.9\totalheight}{\includegraphics[width=3cm]{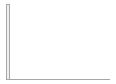}}
&
\begin{itemize}[noitemsep, left=-0.1pt, label={}]
\item 0.0\%
\end{itemize}
\\
\hline
\begin{itemize}[itemsep=4pt, left=0pt, label={}]
\item 29
\end{itemize}
&
\begin{itemize}[itemsep=4pt, left=0pt, label={}]
\item Entropy Buyer Dependence
\item \textnormal{[numeric]}
\end{itemize}
&
\begin{itemize}[noitemsep, left=0pt, label={}]
\item Mean (sd):
\item 0.7 (0.1)
\item min \textless med \textless max:
\item 0 \textless 0.7 \textless 1
\item IQR (CV):
\item 0.1 (0.2)
\end{itemize}
&
\begin{itemize}[noitemsep, left=0pt, label={}]
\item 24502
\item distinct
\item values
\end{itemize}
&
\raisebox{-0.9\totalheight}{\includegraphics[width=3cm]{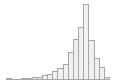}}
&
\begin{itemize}[noitemsep, left=-0.1pt, label={}]
\item 0.0\%
\end{itemize}
\\
\hline
\begin{itemize}[itemsep=4pt, left=0pt, label={}]
\item 30
\end{itemize}
&
\begin{itemize}[itemsep=4pt, left=0pt, label={}]
\item Fragmented Contract Odds
\item \textnormal{[numeric]}
\end{itemize}
&
\begin{itemize}[noitemsep, left=0pt, label={}]
\item Mean (sd):
\item 2.1 (41)
\item min \textless med \textless max:
\item 0 \textless 0.1 \textless 26317.9
\item IQR (CV):
\item 0.6 (19.9)
\end{itemize}
&
\begin{itemize}[noitemsep, left=0pt, label={}]
\item 998666
\item distinct
\item values
\end{itemize}
&
\raisebox{-0.9\totalheight}{\includegraphics[width=3cm]{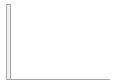}}
&
\begin{itemize}[noitemsep, left=-0.1pt, label={}]
\item 0.0\%
\end{itemize}
\\
\hline
\begin{itemize}[itemsep=4pt, left=0pt, label={}]
\item 31
\end{itemize}
&
\begin{itemize}[itemsep=4pt, left=0pt, label={}]
\item Government level
\item \textnormal{[character]}
\end{itemize}
&
\begin{itemize}[itemsep=4pt, left=0pt, label={}]
\item 1. federal
\item 2. local
\item 3. state
\end{itemize}
&
\begin{itemize}[itemsep=4pt, left=0pt, label={}]
\item 90.5\%
\item 2.8\%
\item 6.7\%
\end{itemize}
&
\raisebox{-0.9\totalheight}{\includegraphics[height=1.5cm]{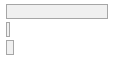}}
&
\begin{itemize}[noitemsep, left=0pt, label={}]
\item 0.0\%
\end{itemize}
\\
\hline
\begin{itemize}[itemsep=4pt, left=0pt, label={}]
\item 32
\end{itemize}
&
\begin{itemize}[itemsep=4pt, left=0pt, label={}]
\item Legal Framework
\item \textnormal{[character]}
\end{itemize}
&
\begin{itemize}[itemsep=4pt, left=0pt, label={}]
\item 1. international
\item 2. national
\item 3. other
\end{itemize}
&
\begin{itemize}[itemsep=4pt, left=0pt, label={}]
\item 17.9\%
\item 81.2\%
\item 0.9\%
\end{itemize}
&
\raisebox{-0.9\totalheight}{\includegraphics[height=1.5cm]{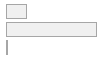}}
&
\begin{itemize}[noitemsep, left=0pt, label={}]
\item 2.9\%
\end{itemize}
\\
\hline
\begin{itemize}[itemsep=4pt, left=0pt, label={}]
\item 33
\end{itemize}
&
\begin{itemize}[itemsep=4pt, left=0pt, label={}]
\item Legal Fundament
\item \textnormal{[character]}
\end{itemize}
&
\begin{itemize}[itemsep=4pt, left=0pt, label={}]
\item 1. art41
\item 2. art42
\item 3. art43
\end{itemize}
&
\begin{itemize}[itemsep=4pt, left=0pt, label={}]
\item 61.8\%
\item 34.4\%
\item 3.8\%
\end{itemize}
&
\raisebox{-0.9\totalheight}{\includegraphics[height=1.5cm]{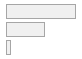}}
&
\begin{itemize}[noitemsep, left=0pt, label={}]
\item 64.5\%
\end{itemize}
\\
\hline
\begin{itemize}[itemsep=4pt, left=0pt, label={}]
\item 34
\end{itemize}
&
\begin{itemize}[itemsep=4pt, left=0pt, label={}]
\item MAD
\item \textnormal{[numeric]}
\end{itemize}
&
\begin{itemize}[noitemsep, left=0pt, label={}]
\item Mean (sd):
\item 0 (0)
\item min \textless med \textless max:
\item 0 \textless 0 \textless 0.2
\item IQR (CV):
\item 0 (0.8)
\end{itemize}
&
\begin{itemize}[noitemsep, left=0pt, label={}]
\item 6823
\item distinct
\item values
\end{itemize}
&
\raisebox{-0.9\totalheight}{\includegraphics[width=3cm]{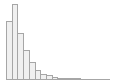}}
&
\begin{itemize}[noitemsep, left=-0.1pt, label={}]
\item 11.4\%
\end{itemize}
\\
\hline
\begin{itemize}[itemsep=4pt, left=0pt, label={}]
\item 35
\end{itemize}
&
\begin{itemize}[itemsep=4pt, left=0pt, label={}]
\item Neighborhood Avg. CRI
\item \textnormal{[numeric]}
\end{itemize}
&
\begin{itemize}[noitemsep, left=0pt, label={}]
\item Mean (sd):
\item 0.5 (0.1)
\item min \textless med \textless max:
\item 0.1 \textless 0.6 \textless 0.9
\item IQR (CV):
\item 0.2 (0.3)
\end{itemize}
&
\begin{itemize}[noitemsep, left=0pt, label={}]
\item 981804
\item distinct
\item values
\end{itemize}
&
\raisebox{-0.9\totalheight}{\includegraphics[width=3cm]{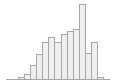}}
&
\begin{itemize}[noitemsep, left=-0.1pt, label={}]
\item 0.1\%
\end{itemize}
\\
\hline
\begin{itemize}[itemsep=4pt, left=0pt, label={}]
\item 36
\end{itemize}
&
\begin{itemize}[itemsep=4pt, left=0pt, label={}]
\item Neighborhood Prop. Recorded-Direct Procedures
\item \textnormal{[numeric]}
\end{itemize}
&
\begin{itemize}[noitemsep, left=0pt, label={}]
\item Mean (sd):
\item 0.7 (0.3)
\item min \textless med \textless max:
\item 0 \textless 0.8 \textless 1
\item IQR (CV):
\item 0.4 (0.4)
\end{itemize}
&
\begin{itemize}[noitemsep, left=0pt, label={}]
\item 266889
\item distinct
\item values
\end{itemize}
&
\raisebox{-0.9\totalheight}{\includegraphics[width=3cm]{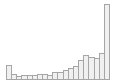}}
&
\begin{itemize}[noitemsep, left=-0.1pt, label={}]
\item 0.1\%
\end{itemize}
\\
\hline
\begin{itemize}[itemsep=4pt, left=0pt, label={}]
\item 37
\end{itemize}
&
\begin{itemize}[itemsep=4pt, left=0pt, label={}]
\item Not in National Registry
\item \textnormal{[integer]}
\end{itemize}
&
\begin{itemize}[itemsep=4pt, left=0pt, label={}]
\item Min: 0
\item Mean: 0.6
\item Max: 1
\end{itemize}
&
\begin{itemize}[itemsep=4pt, left=0pt, label={}]
\item 0: 41.0\%
\item 1: 59.0\%
\end{itemize}
&
\raisebox{-0.9\totalheight}{\includegraphics[height=1.5cm]{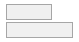}}
&
\begin{itemize}[noitemsep, left=0pt, label={}]
\item 0.0\%
\end{itemize}
\\
\hline
\begin{itemize}[itemsep=4pt, left=0pt, label={}]
\item 38
\end{itemize}
&
\begin{itemize}[itemsep=4pt, left=0pt, label={}]
\item Procedure venue
\item \textnormal{[character]}
\end{itemize}
&
\begin{itemize}[itemsep=4pt, left=0pt, label={}]
\item 1. electronic
\item 2. in-person
\item 3. mixed
\end{itemize}
&
\begin{itemize}[itemsep=4pt, left=0pt, label={}]
\item 34.4\%
\item 45.5\%
\item 20.0\%
\end{itemize}
&
\raisebox{-0.9\totalheight}{\includegraphics[height=1.5cm]{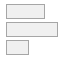}}
&
\begin{itemize}[noitemsep, left=0pt, label={}]
\item 2.9\%
\end{itemize}
\\
\hline
\begin{itemize}[itemsep=4pt, left=0pt, label={}]
\item 39
\end{itemize}
&
\begin{itemize}[itemsep=4pt, left=0pt, label={}]
\item R.F. Benford Law Conformity
\item \textnormal{[numeric]}
\end{itemize}
&
\begin{itemize}[itemsep=4pt, left=0pt, label={}]
\item Mean (sd): 0.7 (0.3)
\item min \textless med \textless max:
\item 0 \textless 1 \textless 1
\item IQR (CV): 0.8 (0.5)
\end{itemize}
&
\begin{itemize}[itemsep=4pt, left=0pt, label={}]
\item 0.00: 6.1\%
\item 0.25: 20.0\%
\item 0.50: 11.4\%
\item 0.75: 11.0\%
\item 1.00: 51.5\%
\end{itemize}
&
\raisebox{-0.9\totalheight}{\includegraphics[height=1.5cm]{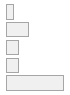}}
&
\begin{itemize}[noitemsep, left=0pt, label={}]
\item 0.0\%
\end{itemize}
\\
\hline
\begin{itemize}[itemsep=4pt, left=0pt, label={}]
\item 40
\end{itemize}
&
\begin{itemize}[itemsep=4pt, left=0pt, label={}]
\item R.F. Buyer Dependence
\item \textnormal{[numeric]}
\end{itemize}
&
\begin{itemize}[noitemsep, left=0pt, label={}]
\item Mean (sd):
\item 0 (0.1)
\item min \textless med \textless max:
\item 0 \textless 0 \textless 1
\item IQR (CV):
\item 0 (3.1)
\end{itemize}
&
\begin{itemize}[noitemsep, left=0pt, label={}]
\item 965094
\item distinct
\item values
\end{itemize}
&
\raisebox{-0.9\totalheight}{\includegraphics[width=3cm]{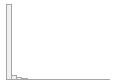}}
&
\begin{itemize}[noitemsep, left=-0.1pt, label={}]
\item 0.0\%
\end{itemize}
\\
\hline
\begin{itemize}[itemsep=4pt, left=0pt, label={}]
\item 41
\end{itemize}
&
\begin{itemize}[itemsep=4pt, left=0pt, label={}]
\item R.F. Decision Period
\item \textnormal{[numeric]}
\end{itemize}
&
\begin{itemize}[itemsep=4pt, left=0pt, label={}]
\item Mean (sd): 0.6 (0.3)
\item min \textless med \textless max:
\item 0 \textless 0.5 \textless 1
\item IQR (CV): 0.8 (0.6)
\end{itemize}
&
\begin{itemize}[itemsep=4pt, left=0pt, label={}]
\item 0.00: 12.9\%
\item 0.25: 13.9\%
\item 0.50: 30.3\%
\item 0.75: 11.1\%
\item 1.00: 31.8\%
\end{itemize}
&
\raisebox{-0.9\totalheight}{\includegraphics[height=1.5cm]{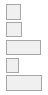}}
&
\begin{itemize}[noitemsep, left=0pt, label={}]
\item 0.0\%
\end{itemize}
\\
\hline
\begin{itemize}[itemsep=4pt, left=0pt, label={}]
\item 42
\end{itemize}
&
\begin{itemize}[itemsep=4pt, left=0pt, label={}]
\item R.F. Procedure Type
\item \textnormal{[numeric]}
\end{itemize}
&
\begin{itemize}[noitemsep, left=0pt, label={}]
\item Mean (sd):
\item 0.8 (0.4)
\item min \textless med \textless max:
\item 0 \textless 1 \textless 1
\item IQR (CV):
\item 0 (0.4)
\end{itemize}
&
\begin{itemize}[noitemsep, left=0pt, label={}]
\item 3
\item distinct
\item values
\end{itemize}
&
\raisebox{-0.9\totalheight}{\includegraphics[width=3cm]{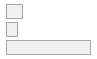}}
&
\begin{itemize}[noitemsep, left=-0.1pt, label={}]
\item 1.0\%
\end{itemize}
\\
\hline
\begin{itemize}[itemsep=4pt, left=0pt, label={}]
\item 43
\end{itemize}
&
\begin{itemize}[itemsep=4pt, left=0pt, label={}]
\item R.F. Single Bidder
\item \textnormal{[character]}
\end{itemize}
&
\begin{itemize}[itemsep=4pt, left=0pt, label={}]
\item 1. Multiple Bidder
\item 2. Single Bidder
\end{itemize}
&
\begin{itemize}[itemsep=4pt, left=0pt, label={}]
\item 89.3\%
\item 10.7\%
\end{itemize}
&
\raisebox{-0.9\totalheight}{\includegraphics[height=1.5cm]{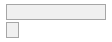}}
&
\begin{itemize}[noitemsep, left=0pt, label={}]
\item 87.9\%
\end{itemize}
\\
\hline
\begin{itemize}[itemsep=4pt, left=0pt, label={}]
\item 44
\end{itemize}
&
\begin{itemize}[itemsep=4pt, left=0pt, label={}]
\item R.F. Submission Period
\item \textnormal{[numeric]}
\end{itemize}
&
\begin{itemize}[itemsep=4pt, left=0pt, label={}]
\item Mean (sd): 0.7 (0.4)
\item min \textless med \textless max:
\item 0 \textless 1 \textless 1
\item IQR (CV): 0.7 (0.5)
\end{itemize}
&
\begin{itemize}[itemsep=4pt, left=0pt, label={}]
\item 0.00: 12.4\%
\item 0.33: 16.3\%
\item 0.66: 18.2\%
\item 1.00: 53.2\%
\end{itemize}
&
\raisebox{-0.9\totalheight}{\includegraphics[height=1.5cm]{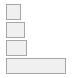}}
&
\begin{itemize}[noitemsep, left=0pt, label={}]
\item 0.0\%
\end{itemize}
\\
\hline
\begin{itemize}[itemsep=4pt, left=0pt, label={}]
\item 45
\end{itemize}
&
\begin{itemize}[itemsep=4pt, left=0pt, label={}]
\item Supplier Avg. CRI
\item \textnormal{[numeric]}
\end{itemize}
&
\begin{itemize}[noitemsep, left=0pt, label={}]
\item Mean (sd):
\item 0.6 (0.2)
\item min \textless med \textless max:
\item 0 \textless 0.6 \textless 1
\item IQR (CV):
\item 0.2 (0.3)
\end{itemize}
&
\begin{itemize}[noitemsep, left=0pt, label={}]
\item 552854
\item distinct
\item values
\end{itemize}
&
\raisebox{-0.9\totalheight}{\includegraphics[width=3cm]{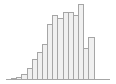}}
&
\begin{itemize}[noitemsep, left=-0.1pt, label={}]
\item 0.0\%
\end{itemize}
\\
\hline
\begin{itemize}[itemsep=4pt, left=0pt, label={}]
\item 46
\end{itemize}
&
\begin{itemize}[itemsep=4pt, left=0pt, label={}]
\item Supplier Betweenness Centrality
\item \textnormal{[numeric]}
\end{itemize}
&
\begin{itemize}[noitemsep, left=0pt, label={}]
\item Mean (sd):
\item 0 (0)
\item min \textless med \textless max:
\item 0 \textless 0 \textless 0.1
\item IQR (CV):
\item 0 (5.7)
\end{itemize}
&
\begin{itemize}[noitemsep, left=0pt, label={}]
\item 105047
\item distinct
\item values
\end{itemize}
&
\raisebox{-0.9\totalheight}{\includegraphics[width=3cm]{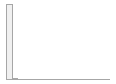}}
&
\begin{itemize}[noitemsep, left=-0.1pt, label={}]
\item 0.0\%
\end{itemize}
\\
\hline
\begin{itemize}[itemsep=4pt, left=0pt, label={}]   
\item 47
\end{itemize}
&
\begin{itemize}[itemsep=4pt, left=0pt, label={}]      
  \item Supplier Closeness Centrality
  \item \textnormal{[numeric]}
\end{itemize}
&
\begin{itemize}[noitemsep, left=0pt, label={}]
  \item Mean (sd):
  \item 0.4 (0.1)
  \item min \textless med \textless max:
  \item 0 \textless 0.4 \textless 0.6
  \item IQR (CV):
  \item 0.1 (0.2)
\end{itemize}
&
\begin{itemize}[noitemsep, left=0pt, label={}] 
  \item 113681
  \item distinct
  \item values      
\end{itemize}
&
\raisebox{-0.9\totalheight}{\includegraphics[width=3cm]{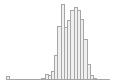}}
&
\begin{itemize}[noitemsep, left=0pt, label={}] 
  \item 0.0\%
\end{itemize}
\\ \hline

\begin{itemize}[itemsep=4pt, left=0pt, label={}]   
\item 48
\end{itemize}
&
\begin{itemize}[itemsep=4pt, left=0pt, label={}]      
  \item Supplier Coreness (avg. degree)
  \item \textnormal{[numeric]}
\end{itemize}
&
\begin{itemize}[noitemsep, left=0pt, label={}]
  \item Mean (sd):
  \item 0.2 (0.3)
  \item min \textless med \textless max:
  \item 0 \textless 0 \textless 1
  \item IQR (CV):
  \item 0.1 (1.9)
\end{itemize}
&
\begin{itemize}[noitemsep, left=0pt, label={}] 
  \item 1309
  \item distinct
  \item values      
\end{itemize}
&
\raisebox{-0.9\totalheight}{\includegraphics[width=3cm]{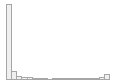}}
&
\begin{itemize}[noitemsep, left=0pt, label={}] 
  \item 0.0\%
\end{itemize}
\\ \hline

\begin{itemize}[itemsep=4pt, left=0pt, label={}]   
\item 49
\end{itemize}
&
\begin{itemize}[itemsep=4pt, left=0pt, label={}]      
  \item Supplier Coreness (strength)
  \item \textnormal{[numeric]}
\end{itemize}
&
\begin{itemize}[noitemsep, left=0pt, label={}]
  \item Mean (sd):
  \item 0.2 (0.3)
  \item min \textless med \textless max:
  \item 0 \textless 0.1 \textless 1
  \item IQR (CV):
  \item 0.4 (1.4)
\end{itemize}
&
\begin{itemize}[noitemsep, left=0pt, label={}] 
  \item 1956
  \item distinct
  \item values      
\end{itemize}
&
\raisebox{-0.9\totalheight}{\includegraphics[width=3cm]{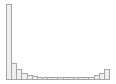}}
&
\begin{itemize}[noitemsep, left=0pt, label={}] 
  \item 0.0\%
\end{itemize}
\\ \hline

\begin{itemize}[itemsep=4pt, left=0pt, label={}]   
\item 50
\end{itemize}
&
\begin{itemize}[itemsep=4pt, left=0pt, label={}]      
  \item Supplier Coreness (weighted degree)
  \item \textnormal{[numeric]}
\end{itemize}
&
\begin{itemize}[noitemsep, left=0pt, label={}]
  \item Mean (sd):
  \item 0.3 (0.3)
  \item min \textless med \textless max:
  \item 0 \textless 0.1 \textless 1
  \item IQR (CV):
  \item 0.4 (1.2)
\end{itemize}
&
\begin{itemize}[noitemsep, left=0pt, label={}] 
  \item 598
  \item distinct
  \item values      
\end{itemize}
&
\raisebox{-0.9\totalheight}{\includegraphics[width=3cm]{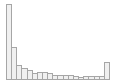}}
&
\begin{itemize}[noitemsep, left=0pt, label={}] 
  \item 0.0\%
\end{itemize}
\\ \hline

\begin{itemize}[itemsep=4pt, left=0pt, label={}]   
\item 51
\end{itemize}
&
\begin{itemize}[itemsep=4pt, left=0pt, label={}]      
  \item Supplier Degree Centrality
  \item \textnormal{[numeric]}
\end{itemize}
&
\begin{itemize}[noitemsep, left=0pt, label={}]
  \item Mean (sd):
  \item 0 (0)
  \item min \textless med \textless max:
  \item 0 \textless 0 \textless 0.5
  \item IQR (CV):
  \item 0 (1.6)
\end{itemize}
&
\begin{itemize}[noitemsep, left=0pt, label={}] 
  \item 24655
  \item distinct
  \item values      
\end{itemize}
&
\raisebox{-0.9\totalheight}{\includegraphics[width=3cm]{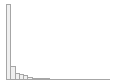}}
&
\begin{itemize}[noitemsep, left=0pt, label={}] 
  \item 0.0\%
\end{itemize}
\\ \hline

\begin{itemize}[itemsep=4pt, left=0pt, label={}]   
\item 52
\end{itemize}
&
\begin{itemize}[itemsep=4pt, left=0pt, label={}]      
  \item Supplier Eigenvector Centrality
  \item \textnormal{[numeric]}
\end{itemize}
&
\begin{itemize}[noitemsep, left=0pt, label={}]
  \item Mean (sd):
  \item 0 (0)
  \item min \textless med \textless max:
  \item 0 \textless 0 \textless 0.2
  \item IQR (CV):
  \item 0 (2.2)
\end{itemize}
&
\begin{itemize}[noitemsep, left=0pt, label={}] 
  \item 192637
  \item distinct
  \item values      
\end{itemize}
&
\raisebox{-0.9\totalheight}{\includegraphics[width=3cm]{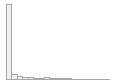}}
&
\begin{itemize}[noitemsep, left=0pt, label={}] 
  \item 0.0\%
\end{itemize}
\\ \hline

\begin{itemize}[itemsep=4pt, left=0pt, label={}]   
\item 53
\end{itemize}
&
\begin{itemize}[itemsep=4pt, left=0pt, label={}]      
  \item Supplier Number of Contracts
  \item \textnormal{[integer]}
\end{itemize}
&
\begin{itemize}[noitemsep, left=0pt, label={}]
  \item Mean (sd):
  \item 153.1 (374.1)
  \item min \textless med \textless max:
  \item 1 \textless 11 \textless 3249
  \item IQR (CV):
  \item 91 (2.4)
\end{itemize}
&
\begin{itemize}[noitemsep, left=0pt, label={}] 
  \item 660
  \item distinct
  \item values      
\end{itemize}
&
\raisebox{-0.9\totalheight}{\includegraphics[width=3cm]{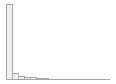}}
&
\begin{itemize}[noitemsep, left=0pt, label={}] 
  \item 0.0\%
\end{itemize}
\\ \hline

\begin{itemize}[itemsep=4pt, left=0pt, label={}]   
\item 54
\end{itemize}
&
\begin{itemize}[itemsep=4pt, left=0pt, label={}]      
  \item Supplier Prop. of Post-Direct Procedures
  \item \textnormal{[numeric]}
\end{itemize}
&
\begin{itemize}[noitemsep, left=0pt, label={}]
  \item Mean (sd):
  \item 0.5 (0.4)
  \item min \textless med \textless max:
  \item 0 \textless 0.5 \textless 1
  \item IQR (CV):
  \item 1 (0.9)
\end{itemize}
&
\begin{itemize}[noitemsep, left=0pt, label={}] 
  \item 2537
  \item distinct
  \item values      
\end{itemize}
&
\raisebox{-0.9\totalheight}{\includegraphics[width=3cm]{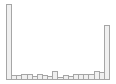}}
&
\begin{itemize}[noitemsep, left=0pt, label={}] 
  \item 0.0\%
\end{itemize}
\\ \hline

\begin{itemize}[itemsep=4pt, left=0pt, label={}]   
\item 55
\end{itemize}
&
\begin{itemize}[itemsep=4pt, left=0pt, label={}]      
  \item Supplier Prop. of Real Direct Procedures
  \item \textnormal{[numeric]}
\end{itemize}
&
\begin{itemize}[noitemsep, left=0pt, label={}]
  \item Mean (sd):
  \item 0.3 (0.4)
  \item min \textless med \textless max:
  \item 0 \textless 0 \textless 1
  \item IQR (CV):
  \item 0.6 (1.5)
\end{itemize}
&
\begin{itemize}[noitemsep, left=0pt, label={}] 
  \item 1760
  \item distinct
  \item values      
\end{itemize}
&
\raisebox{-0.9\totalheight}{\includegraphics[width=3cm]{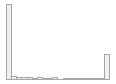}}
&
\begin{itemize}[noitemsep, left=0pt, label={}] 
  \item 0.0\%
\end{itemize}
\\ \hline

\begin{itemize}[itemsep=4pt, left=0pt, label={}]   
\item 56
\end{itemize}
&
\begin{itemize}[itemsep=4pt, left=0pt, label={}]      
  \item Supplier Prop. of Recorded Direct Procedures
  \item \textnormal{[numeric]}
\end{itemize}
&
\begin{itemize}[noitemsep, left=0pt, label={}]
  \item Mean (sd):
  \item 0.7 (0.3)
  \item min \textless med \textless max:
  \item 0 \textless 0.9 \textless 1
  \item IQR (CV):
  \item 0.4 (0.5)
\end{itemize}
&
\begin{itemize}[noitemsep, left=0pt, label={}] 
  \item 2294
  \item distinct
  \item values      
\end{itemize}
&
\raisebox{-0.9\totalheight}{\includegraphics[width=3cm]{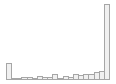}}
&
\begin{itemize}[noitemsep, left=0pt, label={}] 
  \item 0.0\%
\end{itemize}
\\ \hline

\begin{itemize}[itemsep=4pt, left=0pt, label={}]   
\item 57
\end{itemize}
&
\begin{itemize}[itemsep=4pt, left=0pt, label={}]      
  \item Supplier Size
  \item \textnormal{[character]}
\end{itemize}
&
\begin{itemize}[itemsep=4pt, left=0pt, label={}]
  \item 1. medium
  \item 2. micro
  \item 3. not\_mipyme
  \item 4. small
\end{itemize}
&
\begin{itemize}[itemsep=4pt, left=0pt, label={}]
  \item 336137 (20.5\%)
  \item 524840 (32.0\%)
  \item 288145 (17.6\%)
  \item 489807 (29.9\%)
\end{itemize}
&
\raisebox{-0.9\totalheight}{\includegraphics[height=1.5cm]{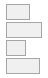}}
&
\begin{itemize}[noitemsep, left=0pt, label={}] 
  \item 669979
  \item (29.0\%)
\end{itemize}
\\ \hline

\begin{itemize}[itemsep=4pt, left=0pt, label={}]   
\item 58
\end{itemize}
&
\begin{itemize}[itemsep=4pt, left=0pt, label={}]      
  \item Supplier Strength
  \item \textnormal{[numeric]}
\end{itemize}
&
\begin{itemize}[noitemsep, left=0pt, label={}]
  \item Mean (sd):
  \item 0 (0)
  \item min \textless med \textless max:
  \item 0 \textless 0 \textless 0
  \item IQR (CV):
  \item 0 (2.1)
\end{itemize}
&
\begin{itemize}[noitemsep, left=0pt, label={}] 
  \item 27433
  \item distinct
  \item values      
\end{itemize}
&
\raisebox{-0.9\totalheight}{\includegraphics[width=3cm]{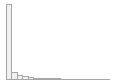}}
&
\begin{itemize}[noitemsep, left=0pt, label={}] 
  \item 0.0\%
\end{itemize}
\\ \hline

\begin{itemize}[itemsep=4pt, left=0pt, label={}]   
\item 59
\end{itemize}
&
\begin{itemize}[itemsep=4pt, left=0pt, label={}]      
  \item Supply Type
  \item \textnormal{[character]}
\end{itemize}
&
\begin{itemize}[itemsep=4pt, left=0pt, label={}]
  \item 1. goods
  \item 2. services
  \item 3. works
\end{itemize}
&
\begin{itemize}[itemsep=4pt, left=0pt, label={}]
  \item 1301679 (56.6\%)
  \item 791044 (34.4\%)
  \item 208684 (9.1\%)
\end{itemize}
&
\raisebox{-0.9\totalheight}{\includegraphics[height=1.5cm]{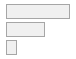}}
&
\begin{itemize}[noitemsep, left=0pt, label={}] 
  \item 7501
  \item (0.3\%)
\end{itemize}
\\ \hline

\begin{itemize}[itemsep=4pt, left=0pt, label={}]   
\item 60
\end{itemize}
&
\begin{itemize}[itemsep=4pt, left=0pt, label={}]      
  \item Tender Period
  \item \textnormal{[numeric]}
\end{itemize}
&
\begin{itemize}[noitemsep, left=0pt, label={}]
  \item Mean (sd):
  \item -15.4 (76.9)
  \item min \textless med \textless max:
  \item -380 \textless 3 \textless 455.4
  \item IQR (CV):
  \item 38 (-5)
\end{itemize}
&
\begin{itemize}[noitemsep, left=0pt, label={}] 
  \item 837
  \item distinct
  \item values      
\end{itemize}
&
\raisebox{-0.9\totalheight}{\includegraphics[width=3cm]{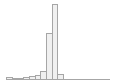}}
&
\begin{itemize}[noitemsep, left=0pt, label={}] 
  \item 401006
  \item (17.4\%)
\end{itemize}
\\ \hline
\caption{Data Frame Summary}
\label{tab:data-summary}
\end{longtable}
}

\newpage
\section{Red Flags and CRI construction}\label{cri-calculation}

Following \cite{fazekas_objective_2016}, the red flags were calculated with a logistic regression using single bidder as dependent variable, Benford Law Conformity, Buyer Dependence (buyer-level variable), Decision Period, Submission Period and Procedure Type as independent variables, and contract year, supply type and region as control variables. The Mexican law admits exceptions to the public procurement process, such as the direct award of contracts, but only under specific circumstances. In our contracts dataset, these contracts are labeled as ``Direct''. However, after a careful review of the contracts, and after contacting Mexican authorities, they confirmed that some contracts categorized as direct procedures, were in reality unsuccessful open procedures. Therefore, we categorized the contract procedures as the following: the Real-Contracts category correspond to the original values of ``direct'' in the contracts dataset. The Post-Direct Contracts category corresponds to the procedures labeled as ``direct'' but that in reality were consequence of failed open procedures. The Real-Direct Contracts are those that initiated as direct procedures.

The significant coefficients of the model were used to rank the corruption risk of the values within each categorical variable. For example, our logistic regression results indicated that within the thresholded variable of decision period, missing values were more likely to be associated with single-bidder contracts, in comparison with the baseline of ``14-365'' days; therefore, we assigned to missing values a value of 1, and to the baseline of a value of 0. Once we get the weights of the red flags, i.e. the ranked corruption risk of the independent variables of the logistic regression, the CRI is calculated simply as the mean of the available red flags for an observation. The CRI is an indicator that will help us to create additional features, but it will not directly used in the models. The exact procedure for creating the red flags is the following:

The general objective is to built the CRI based on several well-known indicators of corruption, such as submission period, decision period, procedure type, buyer dependence, Benford law conformity to fit a logistic regression whose dependent variable is single bidder and controls of contract year, contract price, supply type and region. It is worth to notice that the red flags and CRI were constructed only with a subset of data, were we had information about number of bidders, which correspond to the years 2017-2022. 

\begin{figure}[!ht]
\centering
\includegraphics[width=1\textwidth]{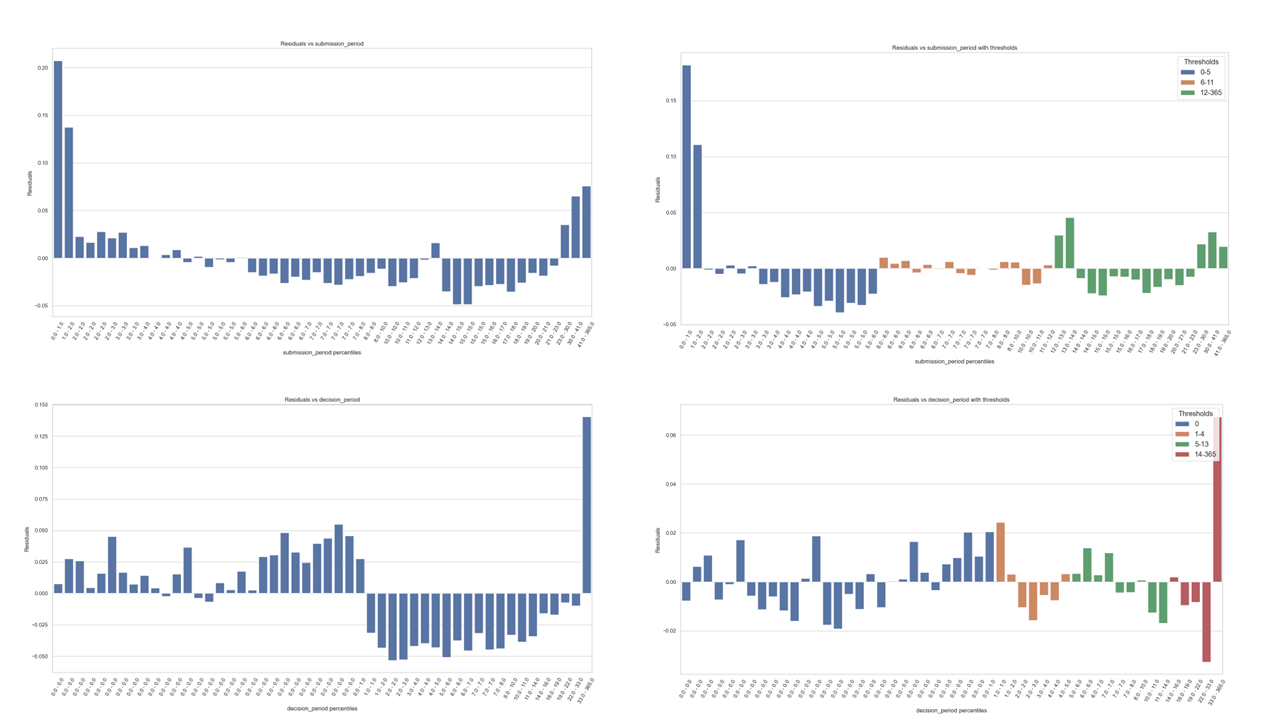}
\caption{Residual Distribution Graphs of Submission Period and Decision Period. This figure shows the distribution of the mean regression residuals of every two percentiles of the variable of interest. Left figures correspond to the un-grouped residuals, and right to the residuals after the variables being thresholded. Top corresponds to submission period and bottom figures to decision period.}\label{fig:residual-analysis}
\end{figure}

Instead of using the numeric values of submission and decision period, these were thresholded using residual distribution graphs. After fitting a logistic regression where the dependent variable is single bidder, the independent variable were the ones to be thresholded (submission and decision period) and the mentioned controls, we plotted the mean regression residuals of every two percentiles of our variable of interest to identify groups of similar underfit or overfit behavior. We then grouped those instances and repeated the process iteratively until the mean regression residuals showed a non-grouping behavior \ref{fig:residual-analysis}. Analyzing the mean regression residuals, we got the thresholds for submission period [``0-5'', ``6-11'', ``12-365''] and [``0'', ``1-4'', ``5-13'', ``14-365''] for decision period. 

For procedure type, the used categories were ``open', ``at-least-three'' and ``post-direct'' procedures. There were no ``real-direct'' procedures in this data subset. For Benford law conformity we used the commonly used definitions of ``close conformity'' [0-0.006), ``acceptable conformity'' [0.006, 0.012), ``marginally acceptable'' [0.012, 0.016) and ``no conformity'' [0.016, inf). For the controls, we used the deciles of contract price instead of the contract price, and region was defined by state. With the purpose of preserve the full population of observations, we also included a missing category to every variable.

We fit single bidder with a logistic regression with procedure type, submission period, decision period, buyer dependence, Benford law conformity as independent variables and  contract year, contract price decile, supply type, and region as controls. We used as baseline the theoretically least level prompt to corruption. The first results indicated that the submission period of ``6-11'' had negative log-odds in comparison with the baseline of ``12-365 days'', therefore, we changed the baseline to the later one. The results of the logistic regression are the following:

\newgeometry{left=3cm,right=3cm,top=2cm,bottom=2cm}

\begin{table}[htbp]
\centering
\begin{tabular}{lclc}
\toprule
\textbf{Dep. Variable:}                                 &  single\_bidder  & \textbf{  No. Observations:  } &   272641    \\
\textbf{Model:}                                         &      Logit       & \textbf{  Df Residuals:      } &   272585    \\
\textbf{Method:}                                        &       MLE        & \textbf{  Df Model:          } &       55    \\
\textbf{  Pseudo R-squ.:}                               &   0.1725         & \textbf{  Log-Likelihood:    } &   -75354.   \\
\textbf{converged:}                                     &       True       & \textbf{  LL-Null:           } &   -91066.   \\
\textbf{Covariance Type:}                               &    nonrobust     & \textbf{  LLR p-value:       } &    0.000    \\
\bottomrule
\end{tabular}
\begin{tabular}{lcccccc}
                                                        & \textbf{coef} & \textbf{std err} & \textbf{z} & \textbf{P$> |$z$|$} & \textbf{[0.025} & \textbf{0.975]}  \\
\midrule
\textbf{Intercept}                                      &      -3.7107  &        0.085     &   -43.471  &         0.000        &       -3.878    &       -3.543     \\
\textbf{procedure\_type\_fixed[T.at\_least\_three]}     &       0.1153  &        0.026     &     4.377  &         0.000        &        0.064    &        0.167     \\
\textbf{procedure\_type\_fixed[T.direct\_failed\_open]} &       1.1466  &        0.056     &    20.467  &         0.000        &        1.037    &        1.256     \\
\textbf{submission\_period\_t[T.12-365]}                &       0.3227  &        0.024     &    13.369  &         0.000        &        0.275    &        0.370     \\
\textbf{submission\_period\_t[T.0-5]}                   &       0.4566  &        0.020     &    23.332  &         0.000        &        0.418    &        0.495     \\
\textbf{submission\_period\_t[T.missing]}               &       1.2828  &        0.077     &    16.677  &         0.000        &        1.132    &        1.434     \\
\textbf{decision\_period\_t[T.0]}                       &       0.7565  &        0.058     &    13.002  &         0.000        &        0.642    &        0.871     \\
\textbf{decision\_period\_t[T.1-4]}                     &       0.5010  &        0.033     &    15.075  &         0.000        &        0.436    &        0.566     \\
\textbf{decision\_period\_t[T.5-13]}                    &       0.1728  &        0.031     &     5.614  &         0.000        &        0.112    &        0.233     \\
\textbf{decision\_period\_t[T.missing]}                 &       0.2017  &        0.095     &     2.121  &         0.034        &        0.015    &        0.388     \\
\textbf{bl\_conformity[T.0\_no\_conformity]}            &       0.9855  &        0.028     &    35.367  &         0.000        &        0.931    &        1.040     \\
\textbf{bl\_conformity[T.1\_marginally\_acceptable]}    &       0.8352  &        0.031     &    26.624  &         0.000        &        0.774    &        0.897     \\
\textbf{bl\_conformity[T.2\_acceptable\_conformity]}    &       0.4013  &        0.026     &    15.592  &         0.000        &        0.351    &        0.452     \\
\textbf{bl\_conformity[T.non\_applicable]}              &       0.7097  &        0.038     &    18.773  &         0.000        &        0.636    &        0.784     \\
\textbf{buyer\_dependence}                              &       1.2900  &        0.088     &    14.704  &         0.000        &        1.118    &        1.462     \\
\bottomrule
\end{tabular}
\caption{Logit Regression Results}\label{logit-single-bidder}
\end{table}
\restoregeometry

\begin{table}[]
\resizebox{\columnwidth}{!}{%
\begin{tabular}{|c|c|c|ccccccc|}
\hline
\textbf{}                                                                             & \multicolumn{1}{l|}{\textbf{}}                                  & \multicolumn{1}{l|}{\textbf{}}                                     & \multicolumn{7}{c|}{\textbf{Assigned risk weight}}                                                                                                                                                                                                                                                                                                                                \\ \hline
\textbf{Red Flag}                                                                     & \textbf{\begin{tabular}[c]{@{}c@{}}Not \\ missing\end{tabular}} & \textbf{\begin{tabular}[c]{@{}c@{}}Missing \\ or NA*\end{tabular}} & \multicolumn{1}{c|}{\textbf{0}} & \multicolumn{1}{c|}{\textbf{0.25}} & \multicolumn{1}{c|}{\textbf{0.33}} & \multicolumn{1}{c|}{\textbf{0.5}}                                              & \multicolumn{1}{c|}{\textbf{0.66}} & \multicolumn{1}{c|}{\textbf{0.75}}                                                   & \textbf{1}                                               \\ \hline
\textbf{\begin{tabular}[c]{@{}c@{}}R.F.  \\ Benford \\ Law\end{tabular}}              & 0.886                                                           & 0.114                                                              & \multicolumn{1}{c|}{close}      & \multicolumn{1}{c|}{acceptable}    & \multicolumn{1}{c|}{}              & \multicolumn{1}{c|}{NA}                                                        & \multicolumn{1}{c|}{}              & \multicolumn{1}{c|}{\begin{tabular}[c]{@{}c@{}}marginally\\ acceptable\end{tabular}} & \begin{tabular}[c]{@{}c@{}}no \\ conformity\end{tabular} \\ \hline
\textbf{\begin{tabular}[c]{@{}c@{}}R.F.   \\ Decision \\ Period\end{tabular}}         & 0.697                                                           & 0.303                                                              & \multicolumn{1}{c|}{14-365}     & \multicolumn{1}{c|}{5-13}          & \multicolumn{1}{c|}{}              & \multicolumn{1}{c|}{missing}                                                   & \multicolumn{1}{c|}{}              & \multicolumn{1}{c|}{1-4}                                                             & 0                                                        \\ \hline
\textbf{\begin{tabular}[c]{@{}c@{}}R.F.   \\ Submission \\ Period \\ **\end{tabular}} & 0.468                                                           & 0.532                                                              & \multicolumn{1}{c|}{6-11}       & \multicolumn{1}{c|}{}              & \multicolumn{1}{c|}{12-365}        & \multicolumn{1}{c|}{}                                                          & \multicolumn{1}{c|}{0-5}           & \multicolumn{1}{c|}{}                                                                & missing                                                  \\ \hline
\textbf{\begin{tabular}[c]{@{}c@{}}R.F.   \\ Single \\ Bidder\end{tabular}}           & 0.121                                                           & 0.893                                                              & \multicolumn{1}{c|}{0}          & \multicolumn{1}{c|}{}              & \multicolumn{1}{c|}{}              & \multicolumn{1}{c|}{}                                                          & \multicolumn{1}{c|}{}              & \multicolumn{1}{c|}{}                                                                & 1                                                        \\ \hline
\textbf{\begin{tabular}[c]{@{}c@{}}R.F.   \\ Procedure \\ Type\end{tabular}}          & 0.99                                                            & 0.01                                                               & \multicolumn{1}{c|}{open}       & \multicolumn{1}{c|}{}              & \multicolumn{1}{c|}{}              & \multicolumn{1}{c|}{\begin{tabular}[c]{@{}c@{}}at \\ least three\end{tabular}} & \multicolumn{1}{c|}{}              & \multicolumn{1}{c|}{}                                                                & direct                                                   \\ \hline
\textbf{\begin{tabular}[c]{@{}c@{}}R.F \\ Buyer \\ Dependence\end{tabular}}           & 1                                                               & 0                                                                  & \multicolumn{1}{c|}{}           & \multicolumn{1}{c|}{}              & \multicolumn{1}{c|}{}              & \multicolumn{1}{c|}{}                                                          & \multicolumn{1}{c|}{}              & \multicolumn{1}{c|}{}                                                                &                                                          \\ \hline
\end{tabular}%
}
\caption{Red Flags thresholds and risk weight. Read exact definitions in \ref{features-definition} \\
* Except for Single Bidder and Procedure Type, all missing or non-applicable were modeled with the other thresholds \\
** The missing values in Submission Period correspond to recorded-direct procedures. 
}\label{red-flags-thresholds}
\end{table}

\FloatBarrier
\clearpage
\section{Details of train-test split}\label{train-test}

\begin{figure}[!ht]
\centering
\includegraphics[width=1\textwidth]{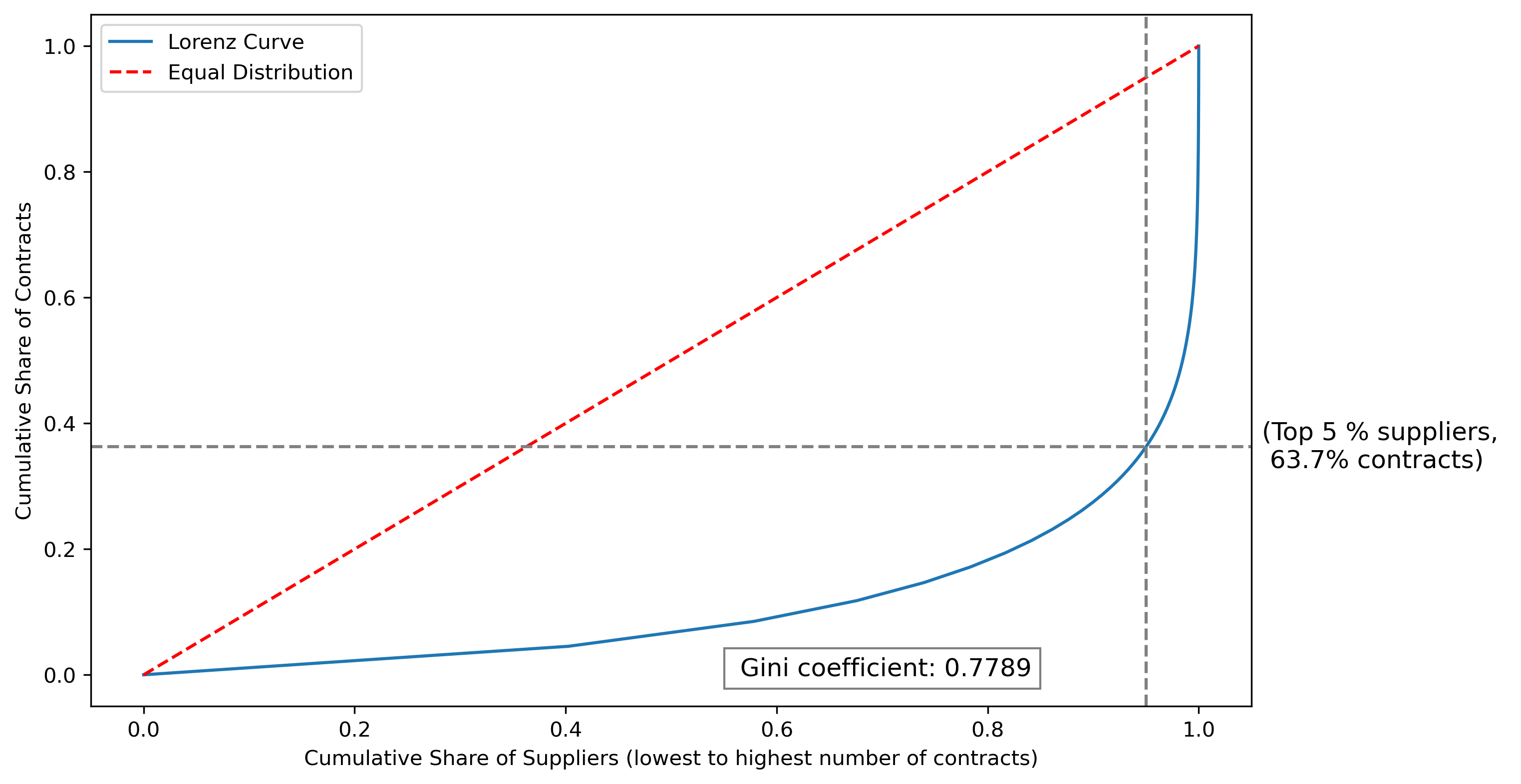}
\caption{Cumulative distribution of contracts among suppliers. The plot shows the cumulative share of contracts ordered by suppliers from the lowest to highest number of contracts. We can observe that the top 5\% accumulates just above 63\% of the contracts, which makes it extremely unequal.}\label{fig:contract-concentration}
\end{figure}

\begin{figure}[!ht]
\centering
\includegraphics[width=\linewidth]{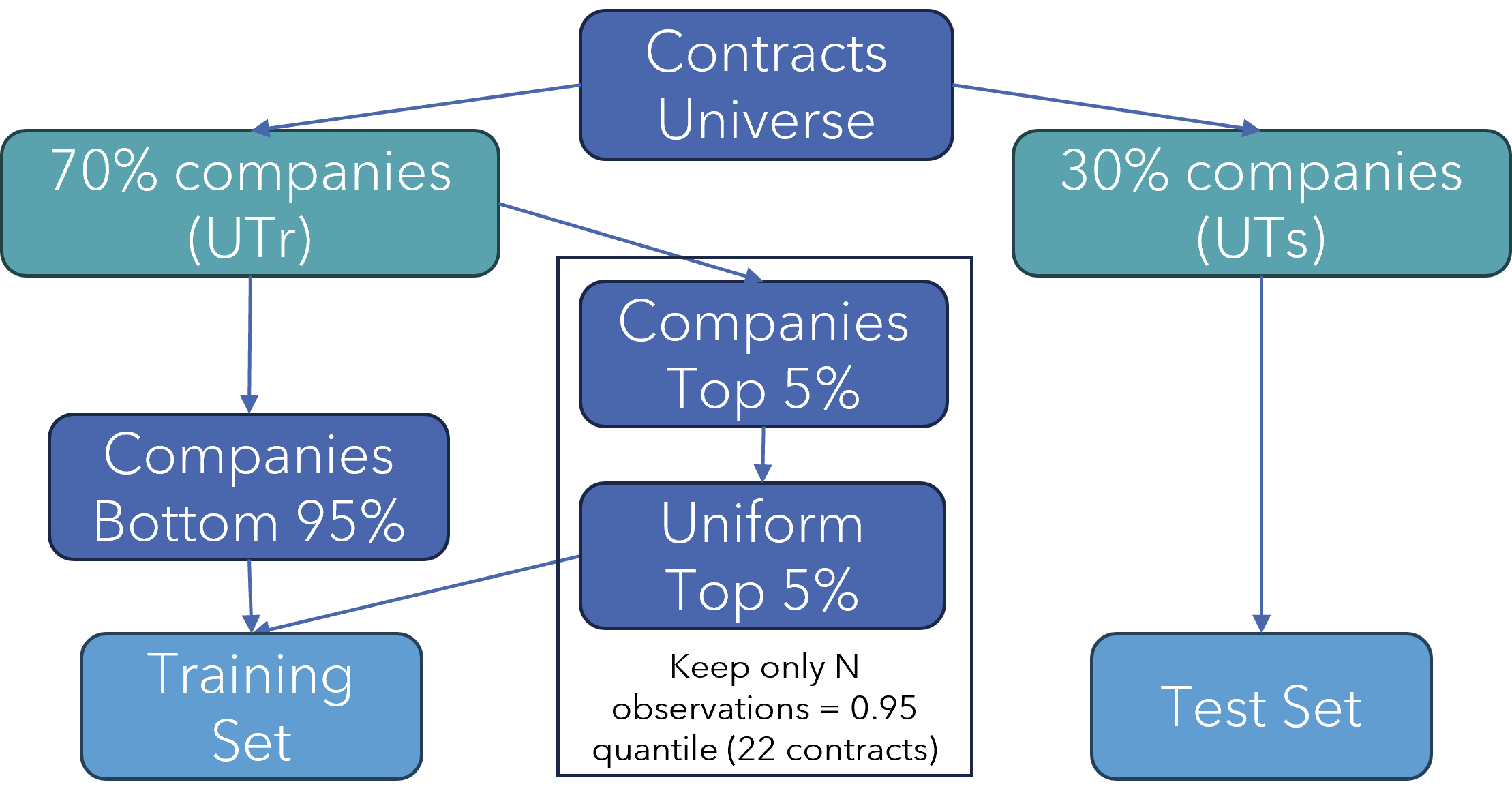}
\caption{Company-based, undersampled, train test split. The figure shows the process of our train-test framework. We first divide the companies on two samples of 70\% and 30\% of the companies. The larger set is subdivided by contracts that belong to the bottom 95\% and the top 5\% of the companies. The top 5\% subset of the companies is undersampled at random so that they only keep at maximum of N contracts corresponding to the the quantile 0.95 of the contract distribution in the first subsample. The bottom 95\% and the test set remain untouched.}\label{fig:train-test}
\end{figure}

\begin{figure}[!ht]
\centering
\includegraphics[width=\linewidth]{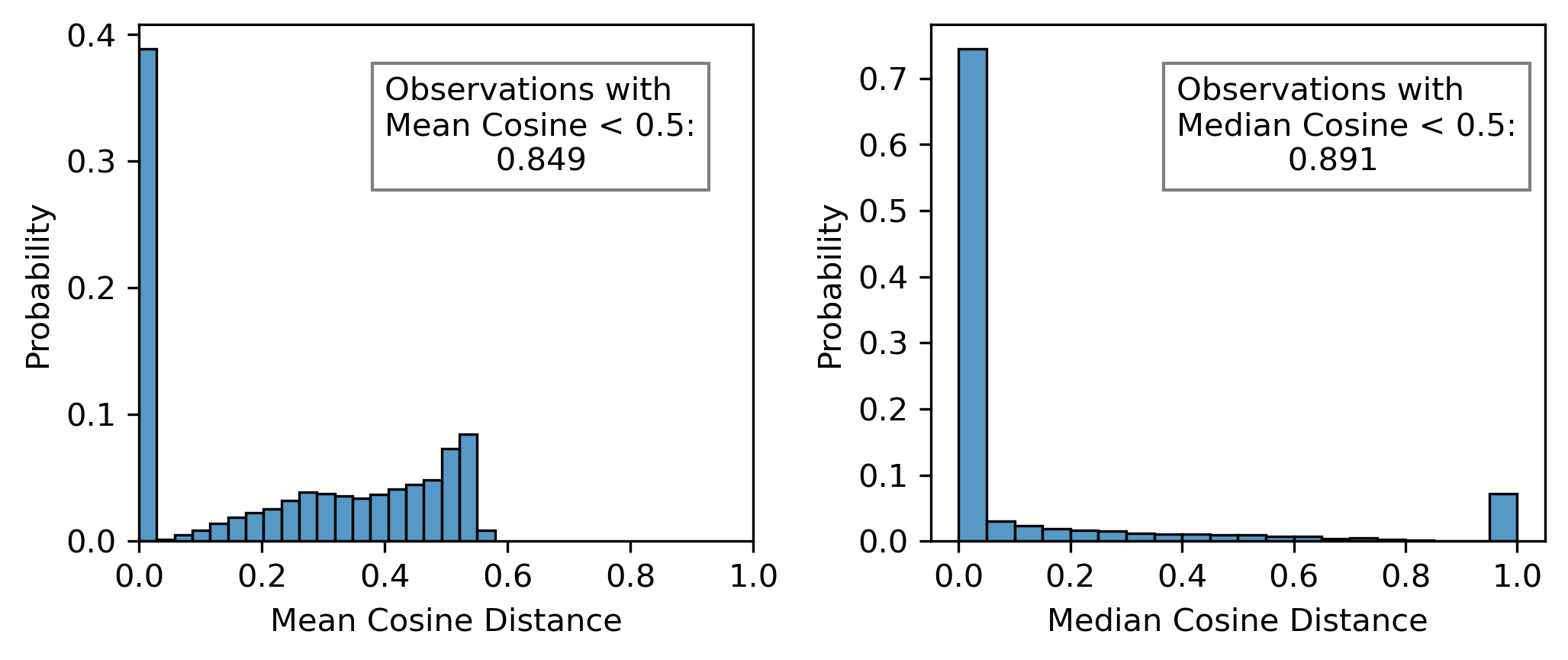}
\caption{Cosine distance distribution of top 5\% suppliers' contracts. The figure show two histograms. On the left one, each observation corresponds to the mean of the cosine distance of a contract with other contracts of the same supplier. On the right one, each observation is the median of the cosine distance of a contract with other contracts of the same supplier. In all cases we only considered to the top 5\% biggest suppliers in terms of their number of contracts. We observe that majority of suppliers have very similar contracts. }\label{fig:cosine_distance_distribution}
\end{figure}

As explained in the main text, the contracts are heavily concentrated in a small number of suppliers \ref{fig:contract-concentration}. In order to prevent the model to learn from a biased sample, we proposed to introduce company-based, undersampled train-test split. The process is the following: we separated our dataset by company, which means, that there's no intersection between contracts of same company between train and test set. In the train test, we then get the top 5\% of companies in terms of number of contracts, and undersample their observations to a maximum that corresponds to the quantile 0.95 of the contract distribution (around 22 contracts) \ref{fig:train-test}. It is reasonable to argue that the undersampling could leave important fraudulent observations out from the training set, however, for each top 5\% supplier, we calculated the cosine distance of all its contracts with themselves to identify if the contracts of a supplier were similar to each other. Our results gave us evidence that the similarity of contracts of same supplier is high (see \ref{fig:cosine_distance_distribution}), therefore we consider that undersampling the contracts of the top 5\% of suppliers is not damaging the representativeness of the sample. 

\FloatBarrier
\clearpage
\section{Hyperparameter selection}

After performing hyperparameter tuning, we identified the optimal configuration for HDSRF as follows:
\begin{itemize}
    \item Class prior: 0.05
    \item Maximum depth: 8
    \item Maximum features: $\sqrt{\text{features}} = 13$
    \item Maximum samples: $K_U$
    \item Minimum samples split: 2
    \item Number of estimators: 1000
\end{itemize}

Due to the complexity of our dataset, when training an SVM learner we employed the Radial Basis Function (RBF) kernel \cite{rahimiRandomFeaturesLargeScale2007} to project the data into a higher-dimensional space, followed by a linear classifier. Specifically, we used stochastic gradient descent (SGD) with hinge loss, which is equivalent to a linear SVM. The RBF kernel hyperparameters were:
\begin{itemize}
    \item $\gamma = 0.01$
    \item Number of components: 200
\end{itemize}

The SGD optimizer was configured with a maximum of 1000 iterations and the hinge loss function. For the PU Bagging framework, we set the number of estimators to 500 and the maximum number of samples to $K_L$.

Here, $K$ denotes the number of instances, with $U$ representing unlabeled instances and $L$ labeled instances. Consequently, HDSRF performs bootstrap sampling with size $U$, whereas PU Bagging constructs balanced sets of positive and unlabeled instances. All remaining parameters were kept at their default values in the \texttt{scikit-learn} framework.

\FloatBarrier
\clearpage
\section{Models' Probability Distributions}\label{probability-distributions}

\begin{figure}[!ht]
\centering
\includegraphics[width=1\textwidth]{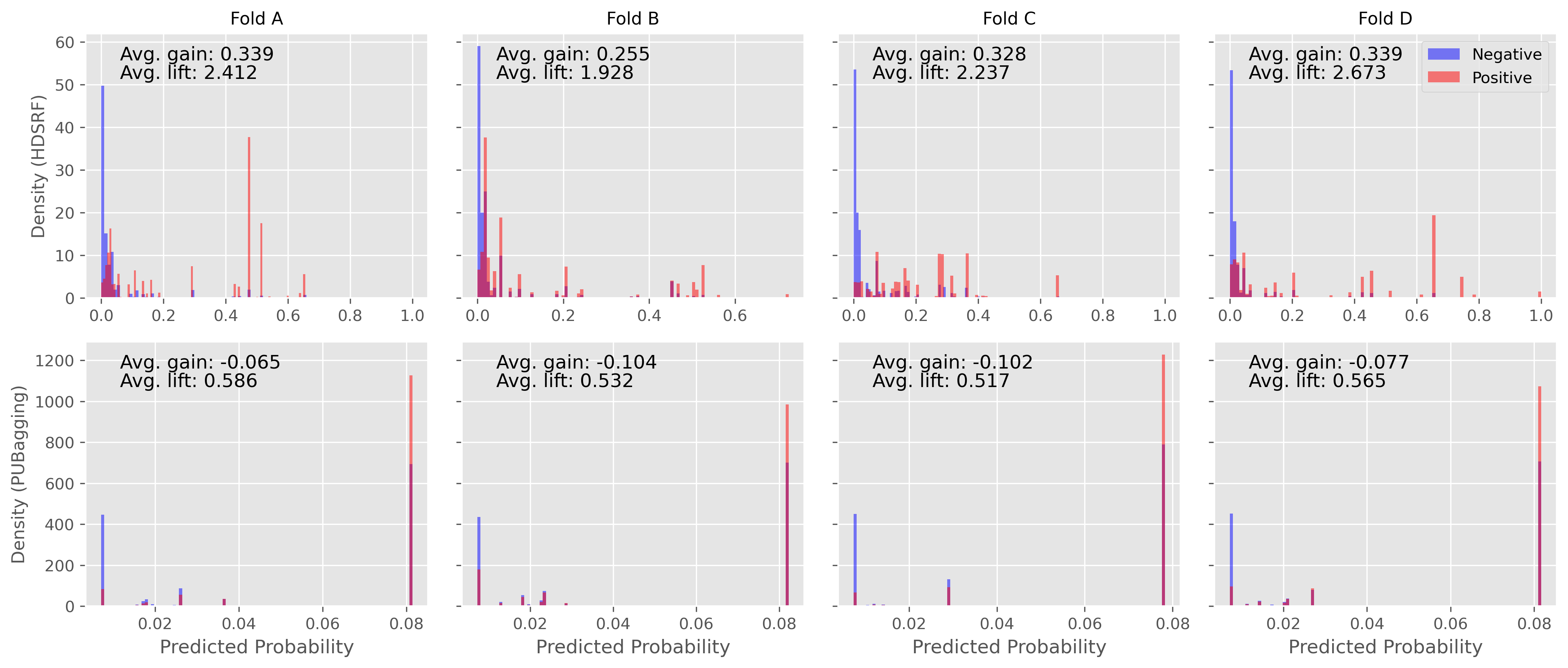}
\caption{Probability Distributions of HDSRF and PUBagging. The figure shows the probability distributions of the HDSRF and PUBagging models with the best performance across 4 cross validation subsets. We observe that the HDSRF tends to have a smoother probability distribution, which makes it more prompt to rank contracts. On the opposite, the PUBagging tends to classify a great number of observations with the same predicted probability.}\label{fig:probabilities}
\end{figure}

\FloatBarrier
\clearpage
\section{Performance on uniform test set}\label{performance-uniform-testset}

\begin{figure}[!ht]
\centering
\includegraphics[width=1\textwidth]{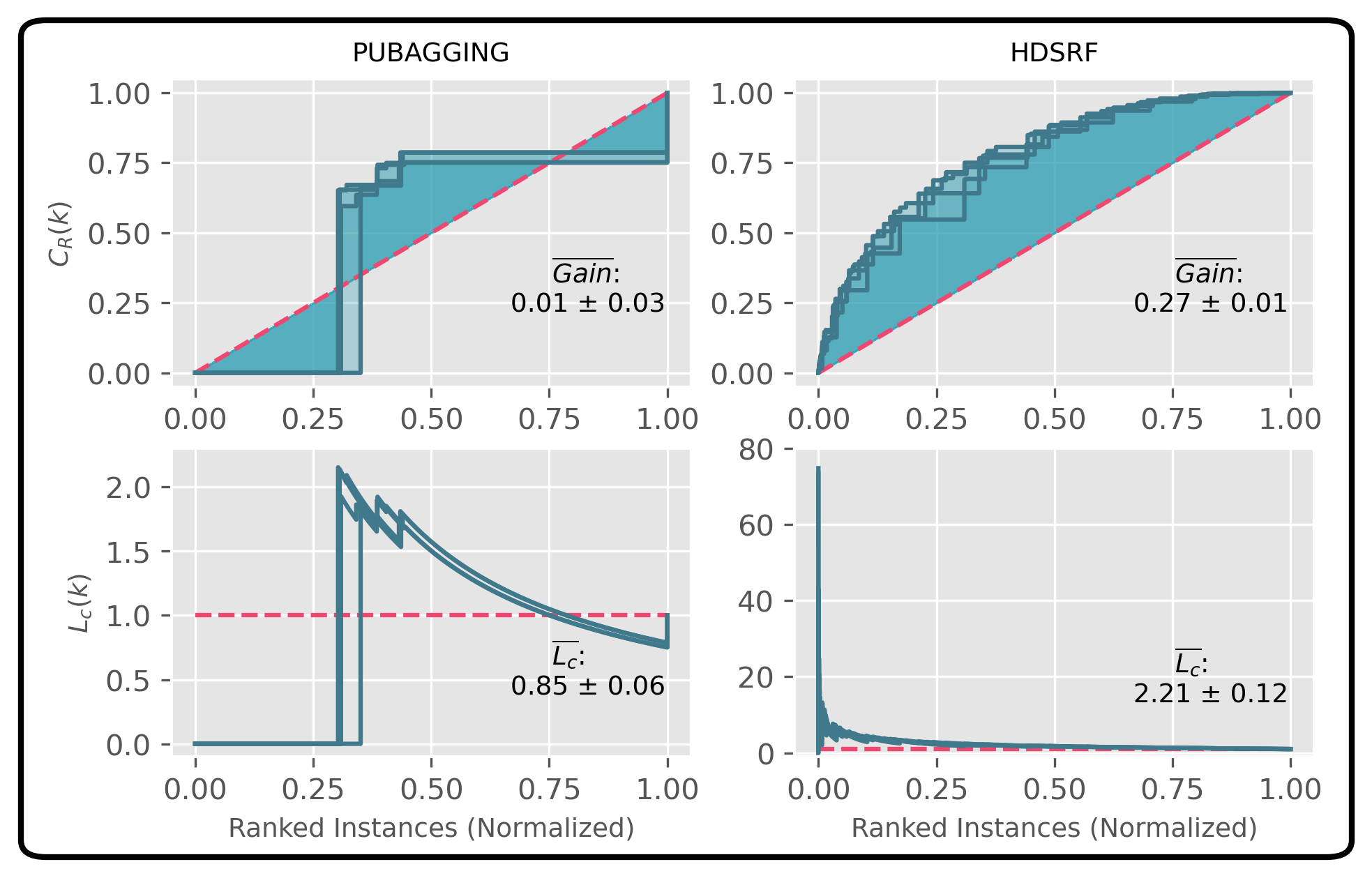}
\caption{Gain and Lift curves. The figure shows the Gain and Lift curves of the examined models in comparison with a random classifier over the cross-validation uniform test-sets. We can observe that the HDSRF outperforms the PU Bagging model in every fold. Moreover, the straight line in the PU Bagging algorithms shows that around 25\% of the observations in the test set are classified with the highest predicted probability, which makes unfeasible to work for ranking purposes.}\label{fig:performance_CSU}
\end{figure}

\FloatBarrier
\clearpage
\section{Performance on inductive setting}\label{performance-inductive}

\begin{figure}[!ht]
\centering
\includegraphics[width=\linewidth]{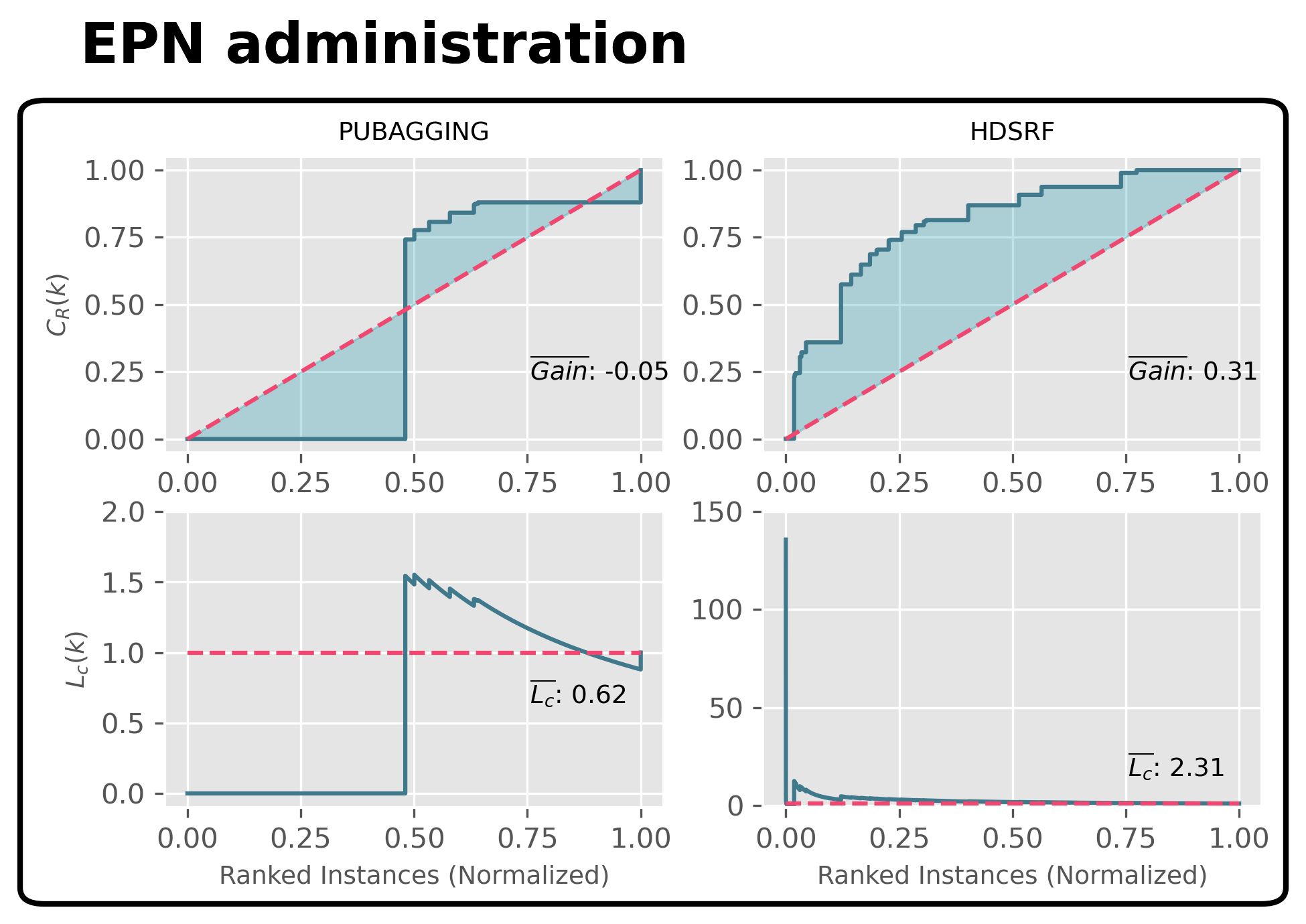}
\caption{Gain and Lift curves for inductive learning in EPN administration. The figure shows the Gain and Lift curves of the examined models in comparison with a random classifier in the inductive setting for the EPN administration. We can observe that the HDSRF outperforms the PU Bagging model greatly, with an average gain of 0.31 and average lift of 2.31, in comparison with the negative average gain and the under performance of average lift in PU Bagging.}\label{fig:ys-epn}
\end{figure}

\begin{figure}[!ht]
\centering
\includegraphics[width=\linewidth]{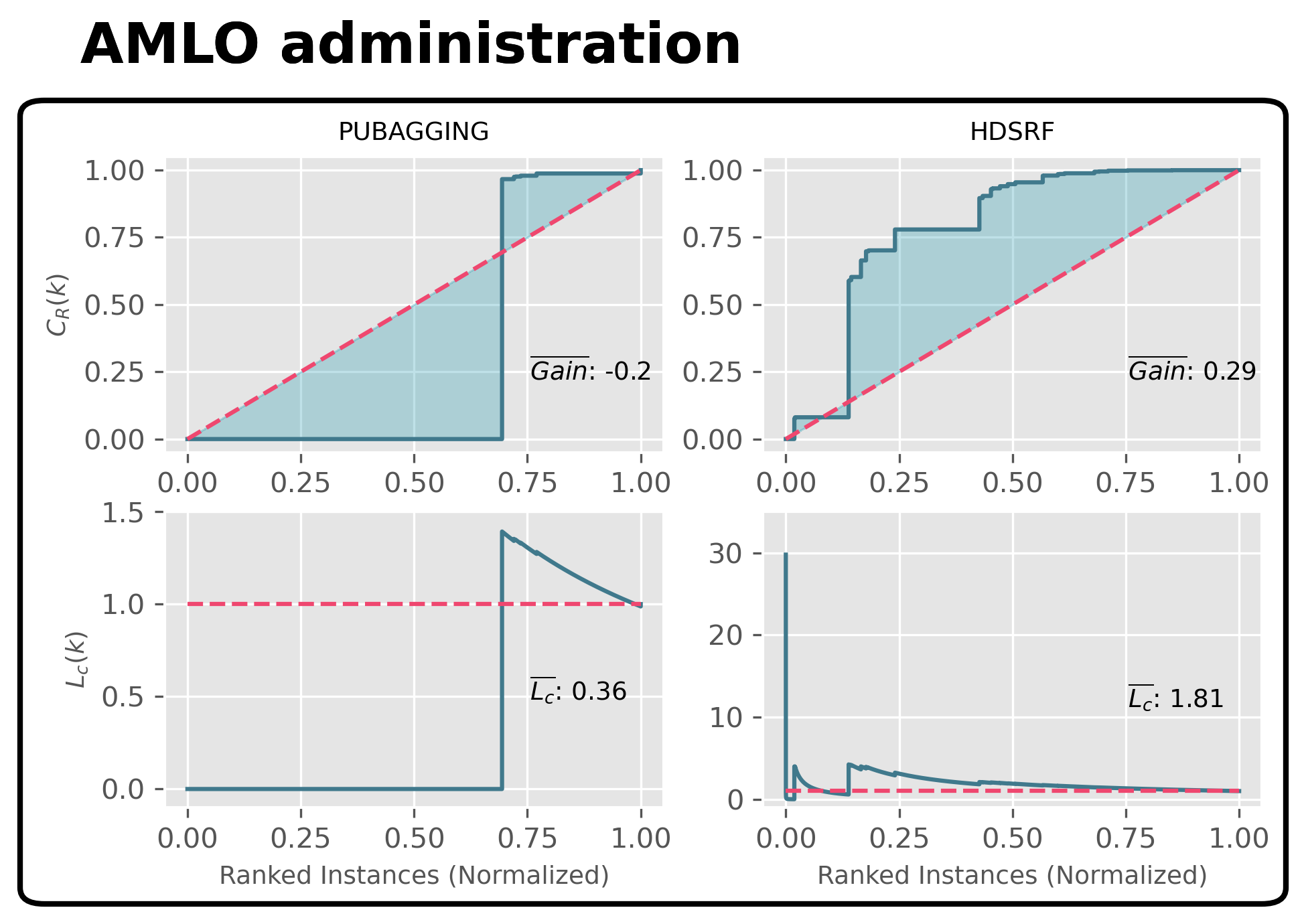}
\caption{Gain and Lift curves for inductive learning in AMLO administration. The figure shows the Gain and Lift curves of the examined models in comparison with a random classifier in the inductive setting for the EPN adminsitration. We can observe that the HDSRF outperforms the PU Bagging model greatly, with an average gain of 0.29 and average lift of 1.81, in comparison with the negative average gain and the under performance of average lift in PU Bagging.}\label{fig:ys-amlo}
\end{figure}

\FloatBarrier
\clearpage
\section{Additional SHAP dependence plots}\label{additional-shap}

\begin{figure}[!ht]
\centering
\includegraphics[width=\linewidth]{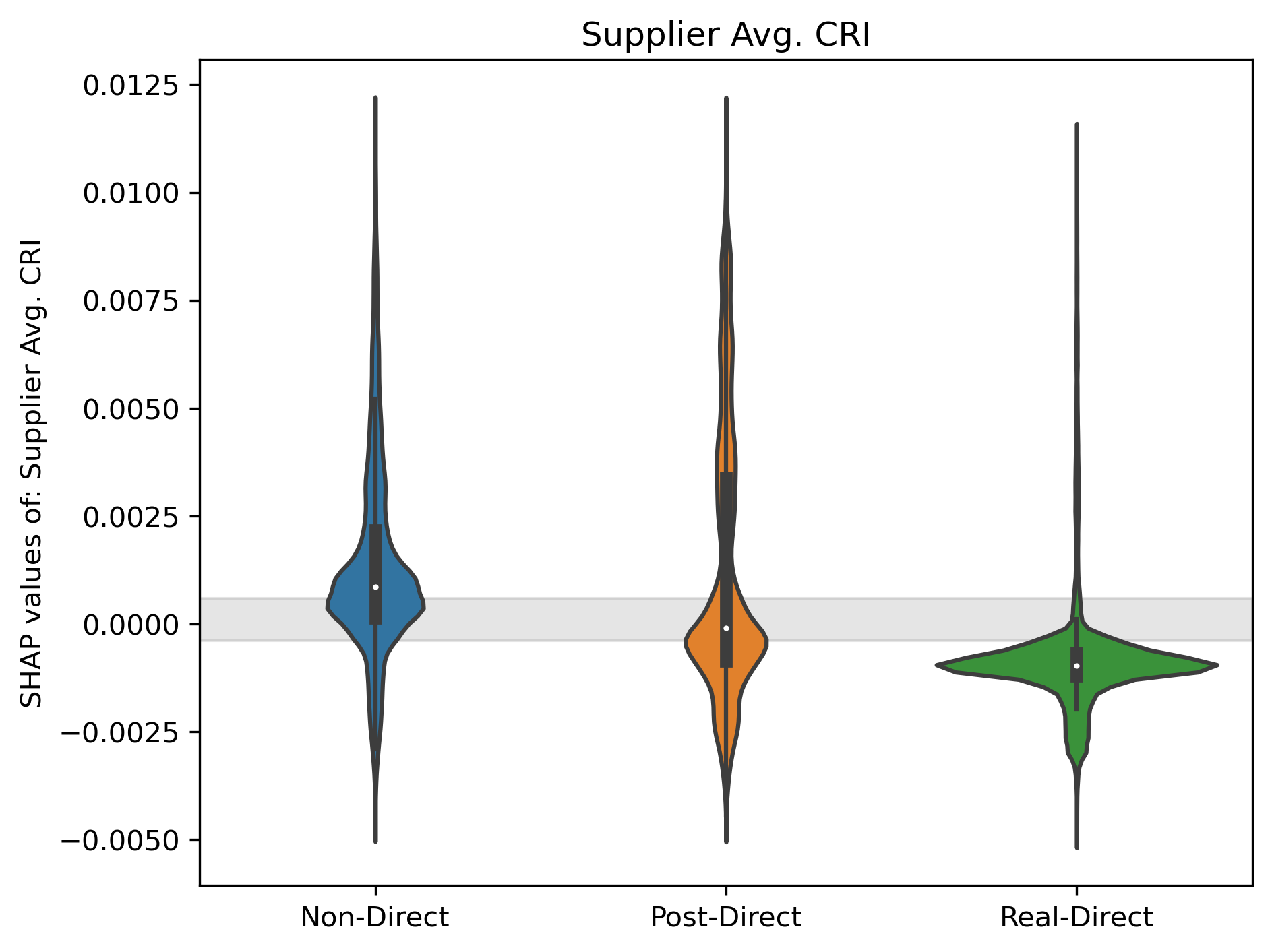}
\caption{SHAP values of Supplier Avg. CRI by procedure type. The figure shows the SHAP values of Avg. CRI by procedure type. We observe that the distribution of the Real-Direct procedures, those that were assigned without competition have the lowest SHAP values. }\label{fig:shap-avgcri}
\end{figure}

\end{appendices}

\end{document}